\documentclass{article} 
\usepackage{iclr2023_conference,times}

\usepackage{amsmath,amsfonts,bm}

\def\eqref#1{equation~\ref{#1}}

\def\1{\bm{1}}

\DeclareMathAlphabet{\mathsfit}{\encodingdefault}{\sfdefault}{m}{sl}
\SetMathAlphabet{\mathsfit}{bold}{\encodingdefault}{\sfdefault}{bx}{n}

\newcommand{\R}{\mathbb{R}}

\usepackage{hyperref}
\usepackage{mathtools}
\usepackage{amssymb}
\usepackage{url}
\usepackage{tabularx}
\usepackage{pdfrender}

\usepackage{microtype}
\usepackage[]{graphicx}
\usepackage{booktabs} 
\usepackage{hyperref}
\usepackage{url}
\usepackage{adjustbox}
\usepackage[toc,page]{appendix}
\usepackage{subcaption}
\usepackage{graphicx,wrapfig,lipsum}
\usepackage{colortbl}
\usepackage{hyperref}

\title{Relational Attention: Generalizing \\ Transformers for Graph-Structured Tasks}

\author{Cameron Diao \thanks{Work was done during an internship at Microsoft Research.} \\
Department of Computer Science \\
Rice University \\
\texttt{cwd2@rice.edu} \\
\And
Ricky Loynd \\
Microsoft Research \\
\texttt{riloynd@microsoft.com} \\
}

\iclrfinalcopy 
\begin{document}

\maketitle

\begin{abstract}
\label{section:abstract}

Transformers flexibly operate over sets of real-valued vectors representing task-specific entities and their attributes, where each vector might encode one word-piece token and its position in a sequence, or some piece of information that carries no position at all.
But as set processors, standard transformers are at a disadvantage in reasoning over more general graph-structured data where nodes represent entities and edges represent relations \textit{between} entities. To address this shortcoming, we generalize transformer attention to consider and update edge vectors in each transformer layer.
We evaluate this \textit{relational transformer} on a diverse array of graph-structured tasks, including the large and challenging CLRS Algorithmic Reasoning Benchmark. There, it dramatically outperforms state-of-the-art graph neural networks expressly designed to reason over graph-structured data.
Our analysis demonstrates that these gains are attributable to relational attention's inherent ability to leverage the greater expressivity of graphs over sets.
\end{abstract}

\section{Introduction}
\label{section:intro}

\begin{wrapfigure}[19]{r}{5cm}
    \centering
    \vspace*{-0.15in}
    \includegraphics[width=\linewidth]{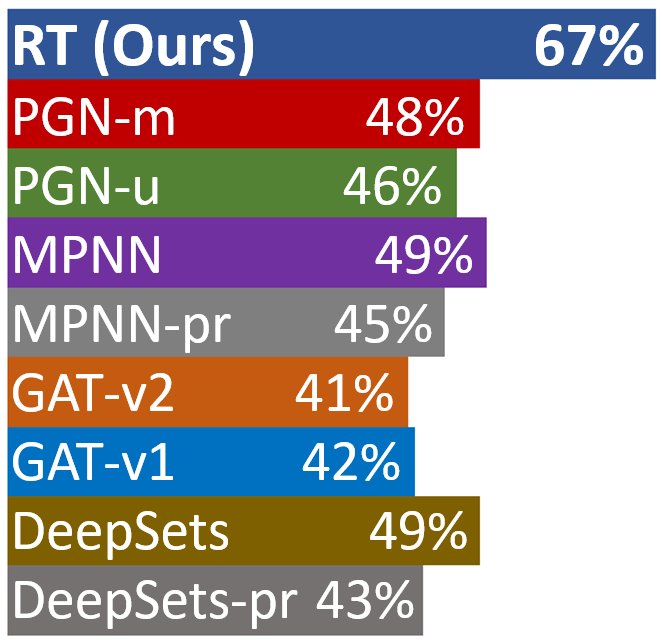}
    \caption{The relational transformer (RT) outperforms baseline GNNs on a set of 30 distinct graph-structured tasks from CLRS-30, averaged by algorithm class.
    }
    \label{fig:bar_chart}
\end{wrapfigure}

Graph-structured problems turn up in many domains, including knowledge bases \citep{WikiKG90M-LS, FB15k}, communication networks \citep{email-EuAll}, citation networks \citep{Cora}, and molecules \citep{MUTAG, NCI1}. One example is predicting the bioactive properties of a molecule, where the atoms of the molecule are the nodes of the graph and the bonds are the edges. Along with their ubiquity, graph-structured problems vary widely in difficulty. For example, certain graph problems can be solved with a simple multi-layer perceptron, while others are quite challenging and require explicit modeling of relational characteristics. 

Graph Neural Networks (GNNs) are designed to process graph-structured data, including the graph's (possibly directed) edge structure and (in some cases) features associated with the edges. In particular, they learn to represent graph features by passing messages between neighboring nodes and edges, and updating the node and (optionally) edge vectors. Importantly, GNNs typically restrict message passing to operate over the edges in the graph.

Standard transformers lack the relational inductive biases \citep{GN} that are explicitly built into the most commonly used GNNs. Instead, the transformer fundamentally consumes \textit{unordered sets} of real-valued vectors, injecting no other assumptions. This allows entities carrying domain-specific attributes (like position) to be encoded as vectors for input to the same transformer architecture applied to different domains. Transformers have produced impressive results in a wide variety of domains, starting with machine translation \citep{VaswaniTransformer}, then quickly impacting language modeling \citep{BERT} and text generation \citep{GPT3}. They are revolutionizing image processing \citep{ViT} and are being applied to a growing variety of settings including reinforcement learning (RL), both online \citep{WMG, GTrXL} and offline \citep{DecisionTransformer, TrajectoryTransformer}. 

Many of the domains transformers succeed in consist of array-structured data, such as text or images. By contrast, graph data is centrally concerned with pairwise relations between entities, represented as edges and edge attributes. Graphs are more general and expressive than sets, in the sense that a set is a special case of a graph---one without edges. So it is not immediately obvious how graph data can be processed by transformers in a way that preserves relational information. Transformers have been successfully applied to graph-structured tasks in one of two broad ways. Certain works, most recently TokenGT \citep{TokenGT}, encode graphs as sets of real-valued vectors passed to a standard transformer. Other works change the transformer architecture itself to consider relational information, e.g. by introducing relative position vectors to transformer attention. We discuss these and many other such approaches in Section \ref{section:related}.

\begin{figure}
    \centering
    \includegraphics[width=1.0\linewidth]{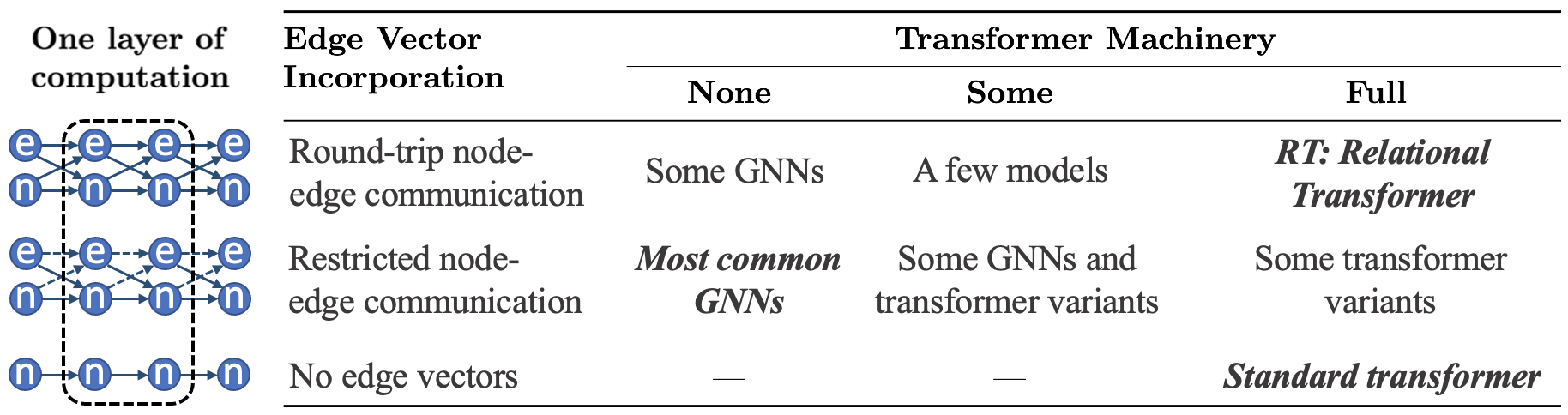}
    \caption{Categories of GNNs and Transformers, compared in terms of transformer machinery and edge vector incorporation. Model categories tested in our experiments are marked in bold.}
    \label{fig:prior_work}
\end{figure}

\textbf{Our novel contribution is relational attention, a mathematically elegant extension of transformer attention, which incorporates edge vectors as first-class model components.}
We call the resulting transformer architecture the Relational Transformer (RT).
As a native graph-to-graph model, RT does not rely on special encoding schemes to input or output graph data.

We find that RT outperforms baseline GNNs on a large and diverse set of difficult graph-structured tasks. In particular, RT establishes dramatically improved state-of-the-art performance (Figure \ref{fig:bar_chart}) over baseline GNNs on the challenging CLRS-30 \citep{CLRS}, which comprises 30 different algorithmic tasks in a framework for probing the reasoning abilities of graph-to-graph models. 
To summarize our main contributions:
\begin{itemize}
  \item We introduce the relational transformer for application to arbitrary graph-structured tasks, and make the implementation available at \url{https://github.com/CameronDiao/relational-transformer}.
  \item We evaluate the reasoning power of RT on a wide range of challenging graph-structured tasks, achieving new state-of-the-art results on CLRS-30.
  \item We enhance the CLRS-30 framework to support evaluation of a broader array of models (Section \ref{subsubsection:fw_enhance}).
  \item We improve the performance of CLRS-30 baseline models by adding multi-layer functionality, and tuning their hyperparameters (Section \ref{subsubsection:model_enhance}).
\end{itemize}


\section{Graph Neural Networks}
\label{section:gnns}

We introduce the graph-to-graph model formalism used in the rest of this paper, inspired by \citet{GN}. The input graph is a directed, attributed graph $G = (\mathcal{N}, \mathcal{E})$, where $\mathcal{N}$ is an unordered set of node vectors $\mathbf{n}_{i} \in \mathbb{R}^{d_{n}}$, $i$ denoting the $i$-th node. $\mathcal{E}$ is a set of edge vectors $\mathbf{e}_{ij} \in \mathbb{R}^{d_{e}}$, where directed edge $(i, j)$ points from node $j$ to node $i$. Each layer $l$ in the model accepts a graph $G^{l}$ as input, processes the graph's features, then outputs graph $G^{l + 1}$ with the same structure as $G^{l}$, but with potentially updated node and edge vectors. Certain tasks may also include a single global vector with the input or output graph, but we omit those details from our formalism since they are not what distinguishes the approaches described below.


Each layer $l$ of a graph-to-graph model is comprised of two update functions $\phi$ and an aggregation function $\bigoplus$,
\begin{align}
    \text{Node vector} \;\; \mathbf{n}^{l + 1}_{i} &= \phi_{n} \left(\mathbf{n}^{l}_{i}, \bigoplus \limits_{j \in \mathcal{L}_{i}} \psi^{m} \left(\mathbf{e}^{l}_{ij}, \mathbf{n}^{l}_{i}, \mathbf{n}^{l}_{j}\right) \right) \label{node update} \\
    \text{Edge vector} \;\; \mathbf{e}^{l + 1}_{ij} &= \phi_{e} \left(\mathbf{e}^{l}_{ij}, \mathbf{e}^{l}_{ji}, \mathbf{n}^{l + 1}_{i}, \mathbf{n}^{l + 1}_{j} \right) \label{edge update}
\end{align}

where 
$\mathcal{L}_{i}$ denotes the set of node $i$'s neighbors (optionally including node $i$), and $\psi^{m}$ denotes a message function.
The baseline GNNs listed below let $\phi_{e}$ be the identity function such that $\mathbf{e}^{l + 1}_{ij} = \mathbf{e}^{l}_{ij}$ for all edges $(i, j)$. Furthermore, they use a permutation-invariant aggregation function for $\bigoplus$. The following details are from \citet{CLRS}:

In \textbf{Deep Sets} \citep{DeepSets}, the only edges are self-connections so each $\mathcal{L}_{i}$ is a singleton set containing node $i$. 

In \textbf{Graph Attention Networks (GAT)} \citep{GAT, GATv2}, $\bigoplus$ is self-attention, and the message function $\psi^{m}$ merely extracts the sender features: $\psi^{m} (\mathbf{e}^{l}_{ij}, \mathbf{n}^{l}_{i}, \mathbf{n}^{l}_{j}) = \mathbf{W}_{m} \mathbf{n}^{l}_{j}$, where $\mathbf{W}_{m}$ is a weight matrix. 

In \textbf{Message Passing Neural Networks (MPNN)} \citep{MPNN}, edges lie between any pair of nodes and $\bigoplus$ is the max pooling operation. 

In \textbf{Pointer Graph Networks (PGN)} \citep{PGN}, edges are constrained by the adjacency matrix and $\bigoplus$ is the max pooling operation. 


\section{Relational Transformer}
\label{section:methods}

We aim to design a mathematically elegant extension of transformer attention, which incorporates edge vectors as first-class model components.
This goal leads us to the following design criteria:
\begin{enumerate}
    \item Preserve all of the transformer's original machinery (though still not fully understood), for its empirically established advantages.
    \item Introduce directed edge vectors to represent relations between entities.
    \item Condition transformer attention on the edge vectors.
    \item Extend the transformer layer to consume edge vectors and produce updated edge vectors.
    \item Preserve the transformer's $\mathcal{O}\left(N^2\right)$ computational complexity.
\end{enumerate}

\subsection{Relational Attention}
\label{subsection:relatt}

(See Appendix \ref{section:appendix_transformer} for a mathematical overview of transformers.)
In addition to accepting node vectors representing entity features (as do all transformers), RT also accepts edge vectors representing relation features, which may include edge-presence flags from an adjacency matrix. But RT operates over a fully connected graph, unconstrained by any input adjacency matrix.

\begin{figure}
    \centering
    \includegraphics[width=0.8\linewidth]{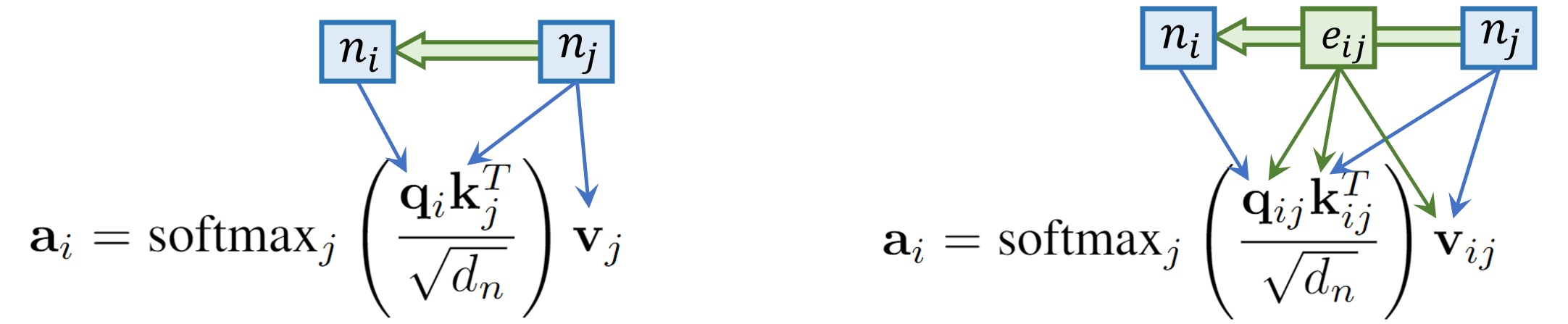}
    \caption{(Left) Standard transformer attention conditions the QKV computation on node vectors. (Right) Relational attention conditions this computation on the intervening edge vector as well.}
    \label{fig:qkv}
\end{figure}

Transformer attention projects QKV vectors from each node vector, 
then computes a dot-product between each pair of vectors $\mathbf{q}_{i}$ and $\mathbf{k}_{j}$.
This dot-product determines the degree to which node $i$ attends to node $j$.
Relational attention's central innovation (illustrated in Figure \ref{fig:qkv}) is to condition the QKV vectors on the directed edge $\mathbf{e}_{ij}$ between the nodes,
by concatenating that edge vector with each node vector prior to the linear transformations:
\begin{equation}
  \mathbf{q}_{ij} = [\mathbf{n}_i, \mathbf{e}_{ij}] {\mathbf{W}^Q} \qquad 
  \mathbf{k}_{ij} = [\mathbf{n}_j, \mathbf{e}_{ij}] {\mathbf{W}^K} \qquad 
  \mathbf{v}_{ij} = [\mathbf{n}_j, \mathbf{e}_{ij}] {\mathbf{W}^V} \qquad
\end{equation}

where each weight matrix $\mathbf{W}$ is now of size $\R^{\left(d_n + d_e\right) {\times} d_n}$, and $d_e$ is the edge vector size.
To implement this efficiently and exactly, we split each weight matrix $\mathbf{W}$ into two separate matrices for projecting node and edge vectors, project the edge vector to three embeddings, then add those to the node's usual attention vectors:
\begin{equation}
  \mathbf{q}_{ij} = \left(\mathbf{n}_i \mathbf{W}_n^Q + \mathbf{e}_{ij} \mathbf{W}_e^Q\right) \qquad 
  \mathbf{k}_{ij} = \left(\mathbf{n}_j \mathbf{W}_n^K + \mathbf{e}_{ij} \mathbf{W}_e^K\right) \qquad 
  \mathbf{v}_{ij} = \left(\mathbf{n}_j \mathbf{W}_n^V + \mathbf{e}_{ij} \mathbf{W}_e^V\right)
\end{equation}

While we have described relational attention in terms of fully connected self-attention, it applies equally to restricted forms of attention such as causal attention, cross-attention, or even restricted GAT-like attention that passes messages only over the edges present in a graph's adjacency matrix.
Relational attention is compatible with multi-head attention, and leaves the transformer's $\mathcal{O}\left(N^2\right)$ complexity unchanged.
Our implementation maintains the high GPU utilization that makes transformers efficient.

\subsection{Edge Updates}
\label{subsection:edgeupdates}

To update edge vectors in each layer of processing,
we follow the general pattern used by transformers to update node vectors: first aggregate messages into one, then use the result to perform a local update.
In self-attention each node attends to all nodes in the graph. 
But having each of the $N^2$ edges attend to $N$ nodes would raise computational complexity to $\mathcal{O}\left(N^3\right)$.
Instead, we restrict the edge's aggregation function to gather messages only from its immediate locale,
which consists of its two adjoining nodes, itself, and the directed edge running in the opposite direction:
\begin{equation}
\mathbf{e}^{l + 1}_{ij} = \phi_{e} \left(\mathbf{e}^{l}_{ij}, \mathbf{e}^{l}_{ji}, \mathbf{n}^{l + 1}_{i}, \mathbf{n}^{l + 1}_{j} \right)
\label{RT edge update}
\end{equation}

We compute the aggregated message $\mathbf{m}^l_{ij}$ by first concatenating the four neighbor vectors, then applying a single linear transformation to the concatenated vector, followed by a ReLU non-linearity:
\begin{equation}
    \mathbf{m}^l_{ij} = \text{ReLU} \left( \text{concat} \left(\mathbf{e}^{l}_{ij}, \mathbf{e}^{l}_{ji}, \mathbf{n}^{l + 1}_{i}, \mathbf{n}^{l + 1}_{j} \right) \mathbf{W}_4 \right)
\end{equation}

where $\mathbf{W}_4 \in \R^{(2d_e + 2d_n) {\times} d_{eh1}}$, and the non-linear ReLU operation takes the place of the non-linear softmax in regular attention.

The remainder of the edge update function is essentially identical to the transformer node update function:
\begin{equation}
    \mathbf{u}^{l}_{ij} = \text{LayerNorm} \left(\mathbf{m}^l_{ij} \mathbf{W}_5 + \mathbf{e}^l_{ij}\right) \quad
    \mathbf{e}^{l+1}_{ij} = \text{LayerNorm} \left(\text{ReLU} (\mathbf{u}^l_{ij} \mathbf{W}_6) \mathbf{W}_7 + \mathbf{u}^l_{ij}\right)
\end{equation}

where $\mathbf{W}_5 \in \R^{d_{eh1} {\times} d_{e}}$,
      $\mathbf{W}_6 \in \R^{d_e {\times} d_{eh2}}$,
      $\mathbf{W}_7 \in \R^{d_{eh2} {\times} d_{e}}$,
and $d_{eh1}$ and $d_{eh2}$ are the hidden layer sizes of the edge feed-forward networks.
Node vectors are updated before edge vectors within each RT layer, so that a node aggregates information from the entire graph
before its adjoining edges use that information in their local updates.
This is another instance of the aggregate-then-update pattern employed by both transformers and GNNs for node vectors.

\section{Prior Work}
\label{section:related}

We categorize, then discuss prior works based on their use of transformer machinery. We further divide each category based on edge vector incorporation, highlighting works that use full node-edge round-tripping, here defined as the process in which edge vectors directly condition node updates and node vectors directly condition edge updates. See Figure \ref{fig:prior_work} for visual comparisons.

\textbf{Encoding Graphs to Sets} For clarity, we define the transformer architecture to exclude any modules that encode inputs to the transformer or decode its outputs. Several prior works have made progress on the challenge of applying standard transformers \citep{VaswaniTransformer} to graph-structured tasks by representing graphs as sets of tokens, e.g. with positional encodings. TokenGT \citep{TokenGT}, for example, treats all nodes and edges of a graph as independent tokens, augmented with token-wise embeddings. Graphormer \citep{Graphormer} and Graph-BERT \citep{Graph-BERT} introduce structural encodings that are applied to each node prior to transformer processing. GraphTrans \citep{GraphTrans} and ReFormer \citep{ReFormer} perform initial convolutions or message passing before the transformer module.

\textbf{Relative Position Vectors} A number of prior works have modified self-attention to implement relative positional encodings \citep{Graphormer, GraphTransformer, Shaw, GREAT, XL}.
To compare these formulations with RT, we expand relational attention's dot-product into four terms as follows:
\begin{equation}
  \mathbf{q}_{ij} \mathbf{k}_{ij}^{\top}=\left([\mathbf{n}_i, \mathbf{e}_{ij}] {\mathbf{W}^Q}\right)\left([\mathbf{n}_j, \mathbf{e}_{ij}] \mathbf{W}^K\right)^{\top}
\end{equation}
\begin{equation}
  =\left(\mathbf{n}_i \mathbf{W}_n^Q + \mathbf{e}_{ij} \mathbf{W}_e^Q\right)\left(\mathbf{n}_j \mathbf{W}_n^K + \mathbf{e}_{ij} \mathbf{W}_e^K\right)^{\top}
\end{equation}
\begin{equation}
  =\left(\mathbf{n}_i \mathbf{W}_n^Q + \mathbf{e}_{ij} \mathbf{W}_e^Q\right)\left( \left(\mathbf{n}_j \mathbf{W}_n^K\right)^{\top} + \left(\mathbf{e}_{ij} \mathbf{W}_e^K\right)^{\top} \right)
\end{equation}
\begin{equation}
  = \mathbf{n}_i \mathbf{W}_n^Q \left(\mathbf{n}_j \mathbf{W}_n^K\right)^{\top}
  + \mathbf{n}_i \mathbf{W}_n^Q \left(\mathbf{e}_{ij} \mathbf{W}_e^K\right)^{\top}
  + \mathbf{e}_{ij} \mathbf{W}_e^Q \left(\mathbf{n}_j \mathbf{W}_n^K\right)^{\top}
  + \mathbf{e}_{ij} \mathbf{W}_e^Q \left(\mathbf{e}_{ij} \mathbf{W}_e^K\right)^{\top}
\end{equation}

The transformer of \citet{VaswaniTransformer} employs only the first term.
The transformer of \citet{Shaw} adds part of the second term, leaving out one weight matrix $\mathbf{W}_e^K$, and GREAT \citep{GREAT} adds the entire third term.
The Transformer-XL \citep{XL} and Graph Transformer of \citet{GraphTransformer} use parts of all four terms, but leave out two of the eight weight matrices $\mathbf{W}_e^Q$ and $\mathbf{W}_e^K$.
Each work above uses edge vectors only for relative positional information.
RT employs all four terms, allows the edge vectors to represent arbitrary information depending on the task,
and updates the edge vectors in each layer of computation.


\textbf{Transformers With Restricted Node-Edge Communication} A few prior graph-to-graph transformers restrict round-trip communication between nodes and edges. EGT \citep{EGT} does a form of node-edge round-tripping but introduces single-scalar bottlenecks (per-edge, per-head). The Graph Transformer of \citet{TheGraphTransformer}, SAT \citep{SAT}, and SAN \citep{SAN} condition node attention coefficients on edge vectors, but they do not explicitly condition node vector updates on edge vectors. GRPE \citep{GRPE} also conditions node attention on edge vectors, e.g. by adding edge vectors to the node value vectors. But the edge vectors themselves are not explicitly updated using node vectors.





\textbf{Highly Modified Transformers} \citet{EdgeTransformer} introduce the Edge Transformer which replaces standard transformer attention with a triangular attention mechanism that takes edge vectors into account and updates the edge vectors in each layer.
This differs from RT in three important respects. 
First, triangular attention is a completely novel form of attention, unlike relational attention which is framed as a natural extension of standard transformer attention.
Second, triangular attention ignores node vectors altogether, and thereby requires node input features and node output predictions to be somehow mapped onto edges.
And third, triangular attention's computational complexity is $\mathcal{O}\left(N^3\right)$ in the number of nodes, unlike RT's relational attention which maintains the $\mathcal{O}\left(N^2\right)$ transformer complexity.
Like Edge Transformer, Nodeformer \citep{Nodeformer} employs a novel form of attention, but in this case with $\mathcal{O}\left(N\right)$ complexity. Nodeformer does not perform node-edge round-tripping, and introduces single-scalar bottlenecks per-edge.
Another transformer, GraphGPS \citep{GPS}, is described by the authors as an MPNN+Transformer hybrid, which does support full node-edge round-tripping. Unlike RT, GraphGPS represents a significant departure from the standard transformer architecture. 

Finally, attentional GNNs, such as GAT \citep{GAT}, GATv2 \citep{GATv2}, Edgeformer \citep{Edgeformer}, kgTransformer \citep{kgTransformer}, Relational Graph Transformer \citep{RGT}, HGT \citep{HGT}, and Simple-HGN \citep{Simple-HGN} aggregate features across neighborhoods based on transformer-style attention coefficients. However, unlike transformers, attentional GNNs only compute attention over input edge vectors, and (except in Edgeformer and kgTransformer) the edge vectors are not updated in each layer. In particular, kgTransformer, Relational Graph Transformer, HGT, and Simple-HGN modify transformer attention to consider hetereogeneous structures in the graph data, such that the model can differentiate between types of nodes and edges.

\textbf{Other GNNs} MPNN \citep{MPNN} is a popular GNN that accepts entire edge vectors as input, as do some other works such as MXMNet \citep{MXMNet} and G-MPNN \citep{G-MPNN}. But apart from EGNN \citep{EGNN} and Censnet \citep{CensNet}, relatively few GNNs update the edge vectors themselves. None of these GNNs use the full transformer machinery, and in general many GNNs are designed for specific settings, such as quantum chemistry.

Unlike any of these prior works, RT preserves all of the original transformer machinery, while adding full bidirectional conditioning of node and edge vector updates.

\section{Experiments}
\label{experiments}

We evaluate RT against common GNNs on the diverse set of graph-structured tasks provided by CLRS-30 \citep{CLRS}, which was designed to measure the reasoning abilities of neural networks. This is a common motivation for tasking neural networks to execute algorithms  \citep{RNN-on-programs, NeuralGPUs, NALU, REINFORCE}. RT outperforms baseline GNNs by wide margins, especially on tasks that require processing of node relations (Section \ref{subsubsection:clrs_analysis}).
We further evaluate RT against GNNs on the end-to-end shortest paths task provided by \citet{IterGNN}, where again RT outperforms the baselines (Section \ref{subsection:itergnn}). Our final experiment (Appendix \ref{section:appendix_sokoban}) evaluates RT against a standard transformer on a reinforcement learning task where \textit{no graph structure is provided}. We find that RT decreases error rates of the RL agent significantly.

\subsection{Step-by-Step Reasoning}
\label{subsection:clrs_bench}



In CLRS-30, each step in a task is framed as a graph-to-graph problem, even for algorithms that may seem unrelated to graphs. To give an example, for list sorting algorithms, each input list element is treated as a separate node and predecessor links are added to order the elements. Task data is organized into task inputs, task outputs, and `hints', which are intermediate inputs and outputs for the intervening steps of an algorithm. Data is comprised of combinations of node, edge, and/or global features, which can be of five possible types: scalars, categoricals, masks, binary masks, or pointers. 

CLRS-30 employs an encode-process-decode framework for model evaluation. Input features are encoded using linear layers, then passed to the model (called a processor) being tested. The model performs one step of computation on the inputs, then it outputs a set of node vectors, which are passed back to the model as input on the next step. On each step, the model's output node vectors are decoded by the framework (using linear layers) and compared with the targets (either hints or final outputs) to compute training losses or evaluation scores. 

Certain CLRS-30 tasks provide a global vector with each input graph. As per CLRS-30 specifications, the baseline GNNs handle global vectors by including them as messages in each update step. They do not propagate global vectors through the steps of the algorithm. RT can use two different methods for handling global vectors and we evaluate both in Section \ref{subsubsection:clrs_ablations}.

\subsubsection{Baseline GNNs}
\label{subsubsection:model_enhance}
We began by reproducing the published results of key baseline models on the eight representative tasks (one per algorithm class) listed in Figure 3 of \citet{CLRS} and in our Table \ref{tab:clrs_pt}. 
For several of the following experiments, we refer to these as the \textit{8 core tasks}.
Our results on these tasks agree closely with the published CLRS-30 results (Table \ref{tab:clrs_repr}). See Appendix \ref{subsubsection:dataset_enhance} for details of our train/test protocol.
We chose not to include Memnet in our experiments given our focus on standard GNNs, and given Memnet's poor performance in the original CLRS-30 experiments. Missing details\footnote{See the final comment in https://github.com/deepmind/clrs/issues/92} made it impossible to reproduce the published GAT results on CLRS-30.

The published CLRS-30 results show sharp drops in out-of-distribution (OOD) performance for all models. For instance, MPNN's average evaluation score drops from 96.63\% on the validation set to 51.02\% on the test set.
We note that small training datasets can induce overfitting even in models that are otherwise capable of generalizing to OOD test sets. 
To mitigate this spurious form of overfitting, we expanded the training datasets by 10x to 10,000 examples generated from the same canonical random seed of 1, and evaluated the effects on the 8 core tasks. 
As shown in Table \ref{tab:clrs_size}, expanding the training set significantly boosts the performance of all baseline models. 
For all of the other experiments in this work, we use the larger training sets of 10,000 examples.

Following \citet{CLRS}, we compute results for two separate PGN models, masked (PGN-m) and unmasked (PGN-u), then select the best result on each task to compute the average shown for the combination PGN-c model (which is called PGN in the CLRS-30 results).
Note therefore that PGN-c does not represent a single model. But it does represent the performance that would be achievable by a PGN model that adaptively learned when to use masking.

We found in early experiments that RT obtained far better results than those of the CLRS-30 baseline GNNs. So to further enhance the performance of the baseline GNNs, we extended them to support multiple layers (rounds of message passing) per algorithmic step, and thoroughly tuned their hyperparameters (see Appendix \ref{section:appendix_hps}).
This significantly improved the baseline results (see Table \ref{tab:clrs_tuning_2}).

See Table \ref{tab:clrs_improve_base} for results comparing CLRS-30 baseline model performances with and without our proposed changes. On certain tasks, baseline score variance increased along with mean scores. For example, on the Jarvis’ March task, tuning raised the score of MPNN from 22.99 ± 3.87\% to 59.31 ± 29.3\%. See Appendix \ref{section:appendix_variance} for detailed analysis of score variance.

\subsubsection{Enhancements of CLRS-30's framework}
\label{subsubsection:fw_enhance}
Most GNNs consume and produce node vectors, and many also consume edge vectors or edge types.
However, relatively few GNNs (and none of the CLRS-30 baseline models) are designed to output edge vectors. 
Because of this, the CLRS-30 framework does not support edge vector outputs from a processor network.
To test models such as RT that have these abilities, we extended the CLRS-30 framework to accept edge and global vectors from the processor at each step, and pass these vectors back to the processor as input on the next step. The framework handles node vectors as usual.

In framing algorithms as graph-to-graph tasks, CLRS-30 relies heavily on what it terms a \textit{node pointer}, which is conceptually equivalent to a directed edge pointing from one node to another. 
Since the CLRS-30 baseline models do not output edge vectors, a decoder in the CLRS-30 framework uses the model's output node vectors to create node pointers. 
But for models like RT that output edge vectors, it is more natural to decode node pointers from those edge vectors alone.
To better support such models, we added a flag to enable this modified behavior in the CLRS-30 framework.

\begin{table*}[tb]
\renewcommand{\arraystretch}{1.0}
\caption{Mean test scores of all tuned models on the eight CLRS-30 algorithm classes.
}
\label{tab:clrs_classes}

\small

\begin{center}
\resizebox{\linewidth}{!}{\begin{tabular}{lrrrrrrrr}\hline
{\textbf{Class}} & {\textbf{Deep Sets}} & {\textbf{GAT-v1}} & {\textbf{GAT-v2}} & {\textbf{MPNN}} & {\textbf{PGN-u}} & {\textbf{PGN-m}} & {\textbf{PGN-c}} & {\textbf{RT (Ours)}} \\ \hline

Divide \& Conquer & 12.29\% & 15.19\% & 14.80\% & 16.14\% & 16.71\% & 51.30\% & 51.30\% & {\textbf{66.52\%}} \\

Dynamic Programming & 68.29\% & 63.88\% & 59.22\% & 68.81\% & 68.56\% & 71.07\% & 71.13\% & {\textbf{83.20\%}}  \\ 

Geometry & 65.47\% & 68.94\% & 67.80\% & 83.03\% & 67.77\% & 66.63\% & 67.78\% & {\textbf{84.55\%}} \\

Graphs & 42.09\% & 52.79\% & 55.53\% & 63.30\% & 59.16\% & 64.36\% & 64.59\% & {\textbf{65.33\%}} \\

Greedy & 77.83\% & 75.75\% & 75.03\% & {\textbf{89.40\%}} & 75.30\% & 76.72\% & 76.72\% & 85.32\% \\

Search & 50.99\% & 35.37\% & 38.04\% & 43.94\% & 50.98\% & 54.21\% & 60.39\% & {\textbf{65.31\%}} \\

Sorting & {\textbf{68.89\%}} & 21.25\% & 17.82\% & 27.12\% & 28.93\% & 2.48\% & 28.93\% & 50.01\% \\

Strings & 2.92\% & 1.36\% & 1.57\% & 2.09\% & 1.61\% & 1.17\% & 1.82\% & {\textbf{32.52\%}} \\

\hline

Average & 48.60\% & 41.82\% & 41.23\% & 49.23\% & 46.13\% & 48.49\% & 52.83\% & {\textbf{66.60\%}} \\

 \hline
\end{tabular}}
\end{center}
\end{table*}

\subsubsection{Main results}
\label{subsubsection:clrs_results}
After tuning hyperparameters for all models (Appendix \ref{section:appendix_hps}), we evaluated RT against the six baseline GNNs on all CLRS-30 tasks, using 20 seeds.
The full results are presented in Table \ref{tab:clrs_meanacc}. 
RT outperforms the top-scoring baseline model (MPNN) by \textbf{11\%} overall.
As bolded in the table, RT scores the highest on \textbf{11 out of 30} tasks.
RT is also the best-performing model on \textbf{6 of 8} algorithmic classes (Table \ref{tab:clrs_classes}), and scores the highest when results are averaged over those classes (Figure \ref{fig:bar_chart}). See Table \ref{tab:clrs_catmapping} for the algorithm-class mappings. For convenience, Figure \ref{fig:bar_chart} includes the prior results (labeled as MPNN-pr and DeepSets-pr) for single-layer GNNs from \citet{CLRS}.
In summary, RT significantly outperforms all baseline GNN models over the CLRS-30 tasks.

\subsubsection{Ablations}
\label{subsubsection:clrs_ablations}

Using only the 8 core tasks (except where noted), we perform several ablations to analyze the factors behind RT's solid performance on CLRS-30.

\textbf{Transformer} - We compare RT to a standard, set-based transformer \citep{VaswaniTransformer} by disabling edge vectors and features in RT. 
Table \ref{tab:clrs_pt} shows that performance collapses by almost 40\% without edge vectors and relational attention, even after re-tuning its hyperparameters.

\textbf{Layers} - The tuned RT uses three layers of computation per algorithmic step. When restricted to a single layer, performance drops drastically (Table \ref{tab:clrs_layers}), even after re-tuning the other hyperparameters. However, single-layer RT still outperforms the top-scoring MPNN by \textbf{10.69\%} on the 8 core tasks, suggesting that relational attention improves expressivity even when restricted to a single layer of computation.

\textbf{Global vector} - Many CLRS-30 tasks provide a global feature vector as input to the processor model. 
We designed RT to handle this global vector by either concatenating it to each input node vector, or by passing it to a dedicated core node \citep{WMG, StarTransformer}.
Hyperparameter tuning chose concatenation instead of the core node option, so concatenation was used in all experiments.
But in this ablation, the core-node method obtained slightly higher test scores on 7 tasks that use global vectors as inputs to the processor (Table \ref{tab:clrs_core}).
The score difference of 0.08\% was marginal, providing no empirical basis to prefer one method of handling the global vector over the other.


\textbf{Node pointer decoding} - We assess impact of the flag we added to the CLRS-30 framework, which can be used to decode node pointers from edge vectors only. Compared to using the original decoding procedure, using the flag improved performance by a small amount (0.30\%) (Table \ref{tab:clrs_decode}).

\textbf{Disabling edge updates} - We disable edge updates in RT such that RT relies solely on relational attention to process input features. Table \ref{tab:clrs_edgeupdate} shows the resulting drop in performance, from {\textbf{81.30\%}} to {\textbf{53.99\%}}. This indicates that edge updates are crucial for RT's learning of relational characteristics in the graph data. As a final note, RT without edge updates still outperforms the transformer by \textbf{11.65\%}, demonstrating the effectiveness of relational attention even without updated edges.

\subsubsection{Algorithmic Analysis}
\label{subsubsection:clrs_analysis}

We investigate the reasoning power of RT based on its test performance on specific algorithmic tasks. We only provide possible explanations here, in line with previous work \citep{CLRS, NeuralExecution, GNNAlignment}. 

\textbf{Underperformance} - The greedy class is one of two where RT is outperformed (by just one other model).
The two greedy tasks, activity selector and task scheduling, require selecting node entities that minimize some metric at each step. For example, in task scheduling, the optimal solution involves repeatedly selecting the task with the smallest processing time. The selection step is aligned with max pooling in GNNs: \citet{NeuralExecution} demonstrate how max pooling aligns with making discrete decisions over neighborhoods. Here, each neighborhood represents a set of candidate entities to be selected from. MPNN, PGN-u, and PGN-m all perform max pooling at each step of message passing. On the other hand, RT performs soft attention pooling, which does not align with the discrete decision-making required to execute greedy algorithms. This may explain RT's underperformance on activity selector and task scheduling, as well as Prim's and Kruskal's.


\textbf{Overperformance} - RT overwhelmingly beats baseline GNNs on dynamic programming (DP) tasks. This is surprising, considering that GNNs have been proven to align well with dynamic programming routines \citep{GNN-DP}. To explain RT's overperformance, we consider 1) edge updates and 2) relational attention. For 1), \citet{Triplet-MPNN} observe that several algorithms in CLRS-30, especially those categorized as DP, require edge-based reasoning---where edges store values, and update those values based on other edges' values. These algorithms do not use node representations in their update functions, yet the baseline GNNs can only learn these update functions using message passing between node representations. On the other hand, RT directly supports edge-based reasoning by representing and updating edges. The hypothesis that RT actually uses this ability is supported by the fact that RT beats baseline GNNs on most of the 6 edge-centric tasks (Find Maximum Subarray, Insertion Sort, Matrix Chain Order, and Optimal BST), though not on the other 2 (Dijkstra and Floyd-Warshall). For 2), recall from Section \ref{subsection:relatt} that relational attention is an extension of standard transformer attention. Standard attention itself is a specific instance of the message-passing function described in \citet{GNN-DP}, which is part of the author's framework for aligning GNNs and DP routines. To see how this is the case, the reader can compare our equation \eqref{eq:standardatt} with equation 1 in \cite{GNN-DP}. From this comparison, and from 1), we expect RT to perform well on both of the edge-centric DP tasks Matrix Chain Order and Optimal BST. We find that RT obtains best scores on both of them, by 5.24\% and 3.66\% respectively.

\subsection{End-To-End Algorithmic Reasoning}
\label{subsection:itergnn}

CLRS-30 evaluates reasoning ability by examining how closely a model can emulate each step of an algorithm. But we may also evaluate reasoning ability by training a model to execute an algorithm end-to-end \citep{GNNAlignment}. We use the task provided by \citet{IterGNN} to evaluate RT in this way. Specifically, we task RT with finding a shortest path distance between two nodes in an undirected lobster graph. The node features are one-hot encoded for source, destination, and remaining. The edge features are binary presence values.


\begin{wrapfigure}[10]{t}{4.5cm}
    \centering
    \vspace*{-0.05in}
    \includegraphics[width=1.0\linewidth]{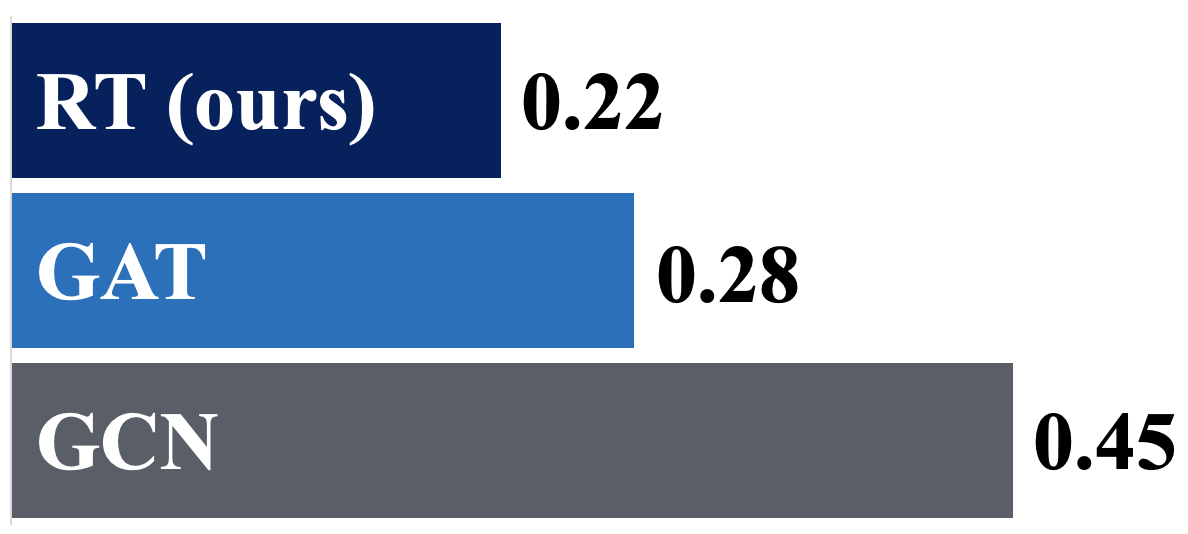}
    \caption{Average Relative Loss on Shortest Paths
    }
    \label{fig:clrs_iter}
\end{wrapfigure}

\textbf{Main Results} We use the same experiment settings as \citet{IterGNN} for the shortest path task. Importantly, we use their method for testing graph size generalizability by training models on graphs of size $[4, 34)$ and evaluating models on graphs of size 100. Furthermore, all models use 30 layers. Results are reported using the relative loss metric introduced by the authors, defined as $\lvert y - \hat{y} \rvert / \lvert y \rvert$ given a label $y$ and a prediction $\hat{y}$. We compare RT to their two baselines, Graph Convolution Network or GCN \citep{GCN} and GAT \citep{GAT}. Results are averaged across 20 random seeds. RT outperforms both baselines with an average relative loss of \textbf{0.22}, compared to GCN's \textbf{0.45} and GAT's \textbf{0.28} (Figure \ref{fig:clrs_iter}). These results were obtained without using the iterative module proposed by \citet{IterGNN} that introduces a stopping criterion to message passing computations.

\section{Conclusion and future work}
\label{section:conc}

We propose the relational transformer (RT), an elegant extension of the standard transformer to operate on graphs instead of sets. It incorporates edge information through relational attention in a principled and computationally efficient way. 
Our experimental results demonstrate that RT performs consistently well across a diverse range of graph-structured tasks.
Specifically, RT outperforms baseline GNNs on CLRS-30 by wide margins, and also outperforms baseline models on end-to-end algorithmic reasoning. 
RT even boosts transformer performance on the Sokoban task, where graph structure is entirely hidden and must be discovered by the RL agent.

Beyond establishing RT's state-of-the-art results on CLRS-30, we enhance performance of the CLRS-30 baseline models, and contribute extensions to the CLRS-30 framework, broadening the scope of models that can be evaluated on the benchmark tasks.
All of these improvements make CLRS-30 tasks and baselines more appealing for evaluating current and future models.

In general, comparing GNNs with transformer-based models like RT on common benchmarks is an important challenge for the community. We have made progress on that challenge by rigorously evaluating RT against standard baseline GNNs on the large and challenging CLRS-30 benchmark, but we leave experiments with other transformer-based approaches for future work. One difficulty is the fact that the CLRS-30 framework is written in Jax, and few if any of these transformers have Jax implementations available. But we have improved the CLRS benchmark itself to make such comparisons more practical in the future. 


In future work, we aim to fully leverage the richness of CLRS-30 to more thoroughly investigate RT's capabilities. For example, recent extensions to the CLRS framework \citep{Triplet-MPNN} allow us to task RT with executing several algorithms simultaneously, which requires knowledge transfer between algorithms. We also plan to evaluate RT on a wider range of real-world graph settings, such as the molecular domain using the large-scale QM9 dataset \citep{MoleculeNet}. Finally, we aim to relax the locality bottleneck of RT's edge updates by allowing edges to attend to other edges directly in a computationally efficient manner.

\subsubsection*{Acknowledgments}

We wish to thank our many collaborators for their valuable feedback, including Roland Fernandez, who also provided the XT ML tool that made research at this scale possible.

\bibliography{iclr2023_conference}

\begin{thebibliography}{60}
\providecommand{\natexlab}[1]{#1}
\providecommand{\url}[1]{\texttt{#1}}
\expandafter\ifx\csname urlstyle\endcsname\relax
  \providecommand{\doi}[1]{doi: #1}\else
  \providecommand{\doi}{doi: \begingroup \urlstyle{rm}\Url}\fi

\bibitem[Battaglia et~al.(2018)Battaglia, Hamrick, Bapst, Sanchez-Gonzalez,
  Zambaldi, Malinowski, Tacchetti, Raposo, Santoro, Faulkner, Gulcehre, Song,
  Ballard, Gilmer, Dahl, Vaswani, Allen, Nash, Langston, Dyer, Heess, Wierstra,
  Kohli, Botvinick, Vinyals, Li, and Pascanu]{GN}
Peter~W. Battaglia, Jessica~B. Hamrick, Victor Bapst, Alvaro Sanchez-Gonzalez,
  Vinicius Zambaldi, Mateusz Malinowski, Andrea Tacchetti, David Raposo, Adam
  Santoro, Ryan Faulkner, Caglar Gulcehre, Francis Song, Andrew Ballard, Justin
  Gilmer, George Dahl, Ashish Vaswani, Kelsey Allen, Charles Nash, Victoria
  Langston, Chris Dyer, Nicolas Heess, Daan Wierstra, Pushmeet Kohli, Matt
  Botvinick, Oriol Vinyals, Yujia Li, and Razvan Pascanu.
\newblock Relational inductive biases, deep learning, and graph networks, 2018.
\newblock URL \url{https://arxiv.org/abs/1806.01261}.

\bibitem[Bergen et~al.(2021)Bergen, O\textquotesingle~Donnell, and
  Bahdanau]{EdgeTransformer}
Leon Bergen, Timothy O\textquotesingle~Donnell, and Dzmitry Bahdanau.
\newblock Systematic generalization with edge transformers.
\newblock In M.~Ranzato, A.~Beygelzimer, Y.~Dauphin, P.S. Liang, and J.~Wortman
  Vaughan (eds.), \emph{Advances in Neural Information Processing Systems},
  volume~34, pp.\  1390--1402. Curran Associates, Inc., 2021.
\newblock URL
  \url{https://proceedings.neurips.cc/paper/2021/file/0a4dc6dae338c9cb08947c07581f77a2-Paper.pdf}.

\bibitem[Bordes et~al.(2013)Bordes, Usunier, Garcia-Dur\'{a}n, Weston, and
  Yakhnenko]{FB15k}
Antoine Bordes, Nicolas Usunier, Alberto Garcia-Dur\'{a}n, Jason Weston, and
  Oksana Yakhnenko.
\newblock Translating embeddings for modeling multi-relational data.
\newblock In \emph{Proceedings of the 26th International Conference on Neural
  Information Processing Systems - Volume 2}, NIPS'13, pp.\  2787–2795, Red
  Hook, NY, USA, 2013. Curran Associates Inc.

\bibitem[Brody et~al.(2022)Brody, Alon, and Yahav]{GATv2}
Shaked Brody, Uri Alon, and Eran Yahav.
\newblock How attentive are graph attention networks?
\newblock In \emph{International Conference on Learning Representations}, 2022.
\newblock URL \url{https://openreview.net/forum?id=F72ximsx7C1}.

\bibitem[Brown et~al.(2020)Brown, Mann, Ryder, Subbiah, Kaplan, Dhariwal,
  Neelakantan, Shyam, Sastry, Askell, Agarwal, Herbert-Voss, Krueger, Henighan,
  Child, Ramesh, Ziegler, Wu, Winter, Hesse, Chen, Sigler, Litwin, Gray, Chess,
  Clark, Berner, McCandlish, Radford, Sutskever, and Amodei]{GPT3}
Tom Brown, Benjamin Mann, Nick Ryder, Melanie Subbiah, Jared~D Kaplan, Prafulla
  Dhariwal, Arvind Neelakantan, Pranav Shyam, Girish Sastry, Amanda Askell,
  Sandhini Agarwal, Ariel Herbert-Voss, Gretchen Krueger, Tom Henighan, Rewon
  Child, Aditya Ramesh, Daniel Ziegler, Jeffrey Wu, Clemens Winter, Chris
  Hesse, Mark Chen, Eric Sigler, Mateusz Litwin, Scott Gray, Benjamin Chess,
  Jack Clark, Christopher Berner, Sam McCandlish, Alec Radford, Ilya Sutskever,
  and Dario Amodei.
\newblock Language models are few-shot learners.
\newblock In H.~Larochelle, M.~Ranzato, R.~Hadsell, M.F. Balcan, and H.~Lin
  (eds.), \emph{Advances in Neural Information Processing Systems}, volume~33,
  pp.\  1877--1901. Curran Associates, Inc., 2020.
\newblock URL
  \url{https://proceedings.neurips.cc/paper/2020/file/1457c0d6bfcb4967418bfb8ac142f64a-Paper.pdf}.

\bibitem[Cai \& Lam(2020{\natexlab{a}})Cai and Lam]{GRPE}
Deng Cai and Wai Lam.
\newblock Graph transformer for graph-to-sequence learning.
\newblock \emph{Proceedings of the AAAI Conference on Artificial Intelligence},
  34:\penalty0 7464--7471, Apr. 2020{\natexlab{a}}.
\newblock \doi{10.1609/aaai.v34i05.6243}.
\newblock URL \url{https://ojs.aaai.org/index.php/AAAI/article/view/6243}.

\bibitem[Cai \& Lam(2020{\natexlab{b}})Cai and Lam]{GraphTransformer}
Deng Cai and Wai Lam.
\newblock Graph transformer for graph-to-sequence learning.
\newblock \emph{Proceedings of the AAAI Conference on Artificial Intelligence},
  34\penalty0 (05):\penalty0 7464--7471, Apr. 2020{\natexlab{b}}.
\newblock \doi{10.1609/aaai.v34i05.6243}.
\newblock URL \url{https://ojs.aaai.org/index.php/AAAI/article/view/6243}.

\bibitem[Chen et~al.(2022)Chen, O'Bray, and Borgwardt]{SAT}
Dexiong Chen, Leslie O'Bray, and Karsten Borgwardt.
\newblock Structure-aware transformer for graph representation learning.
\newblock In \emph{Proceedings of the 39th International Conference on Machine
  Learning~(ICML)}, Proceedings of Machine Learning Research, 2022.

\bibitem[Chen et~al.(2021)Chen, Lu, Rajeswaran, Lee, Grover, Laskin, Abbeel,
  Srinivas, and Mordatch]{DecisionTransformer}
Lili Chen, Kevin Lu, Aravind Rajeswaran, Kimin Lee, Aditya Grover, Misha
  Laskin, Pieter Abbeel, Aravind Srinivas, and Igor Mordatch.
\newblock Decision transformer: Reinforcement learning via sequence modeling.
\newblock In M.~Ranzato, A.~Beygelzimer, Y.~Dauphin, P.S. Liang, and J.~Wortman
  Vaughan (eds.), \emph{Advances in Neural Information Processing Systems},
  volume~34, pp.\  15084--15097. Curran Associates, Inc., 2021.
\newblock URL
  \url{https://proceedings.neurips.cc/paper/2021/file/7f489f642a0ddb10272b5c31057f0663-Paper.pdf}.

\bibitem[Dai et~al.(2019)Dai, Yang, Yang, Carbonell, Le, and Salakhutdinov]{XL}
Zihang Dai, Zhilin Yang, Yiming Yang, Jaime~G. Carbonell, Quoc~Viet Le, and
  Ruslan Salakhutdinov.
\newblock Transformer-xl: Attentive language models beyond a fixed-length
  context.
\newblock In Anna Korhonen, David~R. Traum, and Llu{\'{\i}}s M{\`{a}}rquez
  (eds.), \emph{Proceedings of the 57th Conference of the Association for
  Computational Linguistics, {ACL} 2019, Florence, Italy, July 28- August 2,
  2019, Volume 1: Long Papers}, pp.\  2978--2988. Association for Computational
  Linguistics, 2019.

\bibitem[Debnath et~al.(1991)Debnath, Lopez~de Compadre, Debnath, Shusterman,
  and Hansch]{MUTAG}
Asim~Kumar Debnath, Rosa~L. Lopez~de Compadre, Gargi Debnath, Alan~J.
  Shusterman, and Corwin Hansch.
\newblock Structure-activity relationship of mutagenic aromatic and
  heteroaromatic nitro compounds. correlation with molecular orbital energies
  and hydrophobicity.
\newblock \emph{Journal of Medicinal Chemistry}, 34\penalty0 (2):\penalty0
  786--797, 1991.
\newblock \doi{10.1021/jm00106a046}.
\newblock URL \url{https://doi.org/10.1021/jm00106a046}.

\bibitem[Devlin et~al.(2019)Devlin, Chang, Lee, and Toutanova]{BERT}
Jacob Devlin, Ming-Wei Chang, Kenton Lee, and Kristina Toutanova.
\newblock {BERT}: Pre-training of deep bidirectional transformers for language
  understanding.
\newblock In \emph{Proceedings of the 2019 Conference of the North {A}merican
  Chapter of the Association for Computational Linguistics: Human Language
  Technologies, Volume 1 (Long and Short Papers)}, pp.\  4171--4186,
  Minneapolis, Minnesota, June 2019. Association for Computational Linguistics.
\newblock \doi{10.18653/v1/N19-1423}.
\newblock URL \url{https://aclanthology.org/N19-1423}.

\bibitem[Dosovitskiy et~al.(2021)Dosovitskiy, Beyer, Kolesnikov, Weissenborn,
  Zhai, Unterthiner, Dehghani, Minderer, Heigold, Gelly, Uszkoreit, and
  Houlsby]{ViT}
Alexey Dosovitskiy, Lucas Beyer, Alexander Kolesnikov, Dirk Weissenborn,
  Xiaohua Zhai, Thomas Unterthiner, Mostafa Dehghani, Matthias Minderer, Georg
  Heigold, Sylvain Gelly, Jakob Uszkoreit, and Neil Houlsby.
\newblock An image is worth 16x16 words: Transformers for image recognition at
  scale.
\newblock In \emph{International Conference on Learning Representations}, 2021.
\newblock URL \url{https://openreview.net/forum?id=YicbFdNTTy}.

\bibitem[Dudzik \& Veličković(2022)Dudzik and Veličković]{GNN-DP}
Andrew Dudzik and Petar Veličković.
\newblock Graph neural networks are dynamic programmers, 2022.
\newblock URL \url{https://arxiv.org/abs/2203.15544}.

\bibitem[Dwivedi \& Bresson(2021)Dwivedi and Bresson]{TheGraphTransformer}
Vijay~Prakash Dwivedi and Xavier Bresson.
\newblock A generalization of transformer networks to graphs.
\newblock \emph{AAAI Workshop on Deep Learning on Graphs: Methods and
  Applications}, 2021.

\bibitem[Feng et~al.(2022)Feng, Tan, Li, and Luo]{RGT}
Shangbin Feng, Zhaoxuan Tan, Rui Li, and Minnan Luo.
\newblock Heterogeneity-aware twitter bot detection with relational graph
  transformers.
\newblock \emph{Proceedings of the AAAI Conference on Artificial Intelligence},
  36\penalty0 (4):\penalty0 3977--3985, Jun. 2022.
\newblock \doi{10.1609/aaai.v36i4.20314}.
\newblock URL \url{https://ojs.aaai.org/index.php/AAAI/article/view/20314}.

\bibitem[Gilmer et~al.(2017)Gilmer, Schoenholz, Riley, Vinyals, and Dahl]{MPNN}
Justin Gilmer, Samuel~S. Schoenholz, Patrick~F. Riley, Oriol Vinyals, and
  George~E. Dahl.
\newblock Neural message passing for quantum chemistry.
\newblock In \emph{Proceedings of the 34th International Conference on Machine
  Learning - Volume 70}, ICML'17, pp.\  1263–1272. JMLR.org, 2017.

\bibitem[Gong \& Cheng(2019)Gong and Cheng]{EGNN}
Liyu Gong and Qiang Cheng.
\newblock Exploiting edge features for graph neural networks.
\newblock In \emph{2019 IEEE/CVF Conference on Computer Vision and Pattern
  Recognition (CVPR)}, pp.\  9203--9211, 2019.
\newblock \doi{10.1109/CVPR.2019.00943}.

\bibitem[Guez et~al.(2019)Guez, Mirza, Gregor, Kabra, Racaniere, Weber, Raposo,
  Santoro, Orseau, Eccles, Wayne, Silver, and Lillicrap]{Sokoban}
Arthur Guez, Mehdi Mirza, Karol Gregor, Rishabh Kabra, Sebastien Racaniere,
  Theophane Weber, David Raposo, Adam Santoro, Laurent Orseau, Tom Eccles, Greg
  Wayne, David Silver, and Timothy Lillicrap.
\newblock An investigation of model-free planning.
\newblock In \emph{ICML}, pp.\  2464--2473, 2019.

\bibitem[Guo et~al.(2019)Guo, Qiu, Liu, Shao, Xue, and Zhang]{StarTransformer}
Qipeng Guo, Xipeng Qiu, Pengfei Liu, Yunfan Shao, Xiangyang Xue, and Zheng
  Zhang.
\newblock Star-transformer.
\newblock In \emph{Proceedings of the 2019 Conference of the North {A}merican
  Chapter of the Association for Computational Linguistics: Human Language
  Technologies, Volume 1 (Long and Short Papers)}, pp.\  1315--1325,
  Minneapolis, Minnesota, June 2019. Association for Computational Linguistics.
\newblock \doi{10.18653/v1/N19-1133}.
\newblock URL \url{https://aclanthology.org/N19-1133}.

\bibitem[Hellendoorn et~al.(2020)Hellendoorn, Sutton, Singh, Maniatis, and
  Bieber]{GREAT}
Vincent~J. Hellendoorn, Charles Sutton, Rishabh Singh, Petros Maniatis, and
  David Bieber.
\newblock Global relational models of source code.
\newblock In \emph{International Conference on Learning Representations}, 2020.
\newblock URL \url{https://openreview.net/forum?id=B1lnbRNtwr}.

\bibitem[Hu et~al.(2021)Hu, Fey, Ren, Nakata, Dong, and Leskovec]{WikiKG90M-LS}
Weihua Hu, Matthias Fey, Hongyu Ren, Maho Nakata, Yuxiao Dong, and Jure
  Leskovec.
\newblock Ogb-lsc: A large-scale challenge for machine learning on graphs,
  2021.
\newblock URL \url{https://arxiv.org/abs/2103.09430}.

\bibitem[Hu et~al.(2020)Hu, Dong, Wang, and Sun]{HGT}
Ziniu Hu, Yuxiao Dong, Kuansan Wang, and Yizhou Sun.
\newblock Heterogeneous graph transformer.
\newblock In \emph{Proceedings of The Web Conference 2020}, WWW '20, pp.\
  2704–2710, New York, NY, USA, 2020. Association for Computing Machinery.
\newblock ISBN 9781450370233.
\newblock \doi{10.1145/3366423.3380027}.
\newblock URL \url{https://doi.org/10.1145/3366423.3380027}.

\bibitem[Hussain et~al.(2022)Hussain, Zaki, and Subramanian]{EGT}
Md~Shamim Hussain, Mohammed~J. Zaki, and Dharmashankar Subramanian.
\newblock Global self-attention as a replacement for graph convolution.
\newblock In \emph{Proceedings of the 28th {ACM} {SIGKDD} Conference on
  Knowledge Discovery and Data Mining}. {ACM}, aug 2022.
\newblock \doi{10.1145/3534678.3539296}.
\newblock URL \url{https://doi.org/10.1145%2F3534678.3539296}.

\bibitem[Ibarz et~al.(2022)Ibarz, Kurin, Papamakarios, Nikiforou, Bennani,
  Csord{\'a}s, Dudzik, Bo{\v{s}}njak, Vitvitskyi, Rubanova, Deac, Bevilacqua,
  Ganin, Blundell, and Veli{\v{c}}kovi{\'c}]{Triplet-MPNN}
Borja Ibarz, Vitaly Kurin, George Papamakarios, Kyriacos Nikiforou, Mehdi
  Bennani, R{\'o}bert Csord{\'a}s, Andrew~Joseph Dudzik, Matko Bo{\v{s}}njak,
  Alex Vitvitskyi, Yulia Rubanova, Andreea Deac, Beatrice Bevilacqua, Yaroslav
  Ganin, Charles Blundell, and Petar Veli{\v{c}}kovi{\'c}.
\newblock A generalist neural algorithmic learner.
\newblock In \emph{The First Learning on Graphs Conference}, 2022.
\newblock URL \url{https://openreview.net/forum?id=FebadKZf6Gd}.

\bibitem[Janner et~al.(2021)Janner, Li, and Levine]{TrajectoryTransformer}
Michael Janner, Qiyang Li, and Sergey Levine.
\newblock Offline reinforcement learning as one big sequence modeling problem.
\newblock In M.~Ranzato, A.~Beygelzimer, Y.~Dauphin, P.S. Liang, and J.~Wortman
  Vaughan (eds.), \emph{Advances in Neural Information Processing Systems},
  volume~34, pp.\  1273--1286. Curran Associates, Inc., 2021.
\newblock URL
  \url{https://proceedings.neurips.cc/paper/2021/file/099fe6b0b444c23836c4a5d07346082b-Paper.pdf}.

\bibitem[Jiang et~al.(2019)Jiang, Ji, and Li]{CensNet}
Xiaodong Jiang, Pengsheng Ji, and Sheng Li.
\newblock Censnet: Convolution with edge-node switching in graph neural
  networks.
\newblock In \emph{Proceedings of the Twenty-Eighth International Joint
  Conference on Artificial Intelligence, {IJCAI-19}}, pp.\  2656--2662.
  International Joint Conferences on Artificial Intelligence Organization, 7
  2019.
\newblock \doi{10.24963/ijcai.2019/369}.
\newblock URL \url{https://doi.org/10.24963/ijcai.2019/369}.

\bibitem[Jin et~al.(2023)Jin, Zhang, Meng, and Han]{Edgeformer}
Bowen Jin, Yu~Zhang, Yu~Meng, and Jiawei Han.
\newblock Edgeformers: Graph-empowered transformers for representation learning
  on textual-edge networks.
\newblock In \emph{The Eleventh International Conference on Learning
  Representations}, 2023.
\newblock URL \url{https://openreview.net/forum?id=2YQrqe4RNv}.

\bibitem[Kaiser \& Sutskever(2015)Kaiser and Sutskever]{NeuralGPUs}
Łukasz Kaiser and Ilya Sutskever.
\newblock Neural gpus learn algorithms, 2015.
\newblock URL \url{https://arxiv.org/abs/1511.08228}.

\bibitem[Kim et~al.(2022)Kim, Nguyen, Min, Cho, Lee, Lee, and Hong]{TokenGT}
Jinwoo Kim, Dat~Tien Nguyen, Seonwoo Min, Sungjun Cho, Moontae Lee, Honglak
  Lee, and Seunghoon Hong.
\newblock Pure transformers are powerful graph learners.
\newblock In Alice~H. Oh, Alekh Agarwal, Danielle Belgrave, and Kyunghyun Cho
  (eds.), \emph{Advances in Neural Information Processing Systems}, 2022.
\newblock URL \url{https://openreview.net/forum?id=um2BxfgkT2_}.

\bibitem[Kipf \& Welling(2017)Kipf and Welling]{GCN}
Thomas~N. Kipf and Max Welling.
\newblock Semi-supervised classification with graph convolutional networks.
\newblock In \emph{International Conference on Learning Representations}, 2017.
\newblock URL \url{https://openreview.net/forum?id=SJU4ayYgl}.

\bibitem[Kool et~al.(2019)Kool, van Hoof, and Welling]{REINFORCE}
Wouter Kool, Herke van Hoof, and Max Welling.
\newblock Attention, learn to solve routing problems!
\newblock In \emph{International Conference on Learning Representations}, 2019.
\newblock URL \url{https://openreview.net/forum?id=ByxBFsRqYm}.

\bibitem[Kreuzer et~al.(2021)Kreuzer, Beaini, Hamilton, L{\'e}tourneau, and
  Tossou]{SAN}
Devin Kreuzer, Dominique Beaini, William~L. Hamilton, Vincent L{\'e}tourneau,
  and Prudencio Tossou.
\newblock Rethinking graph transformers with spectral attention.
\newblock In A.~Beygelzimer, Y.~Dauphin, P.~Liang, and J.~Wortman Vaughan
  (eds.), \emph{Advances in Neural Information Processing Systems}, 2021.
\newblock URL \url{https://openreview.net/forum?id=huAdB-Tj4yG}.

\bibitem[Leskovec et~al.(2010)Leskovec, Huttenlocher, and
  Kleinberg]{email-EuAll}
Jure Leskovec, Daniel Huttenlocher, and Jon Kleinberg.
\newblock Signed networks in social media.
\newblock In \emph{Proceedings of the SIGCHI Conference on Human Factors in
  Computing Systems}, CHI '10, pp.\  1361–1370, New York, NY, USA, 2010.
  Association for Computing Machinery.
\newblock ISBN 9781605589299.
\newblock \doi{10.1145/1753326.1753532}.
\newblock URL \url{https://doi.org/10.1145/1753326.1753532}.

\bibitem[Liu et~al.(2022)Liu, Zhao, Su, Cen, Qiu, Zhang, Wu, Dong, and
  Tang]{kgTransformer}
Xiao Liu, Shiyu Zhao, Kai Su, Yukuo Cen, Jiezhong Qiu, Mengdi Zhang, Wei Wu,
  Yuxiao Dong, and Jie Tang.
\newblock Mask and reason: Pre-training knowledge graph transformers for
  complex logical queries.
\newblock In \emph{Proceedings of the 28th {ACM} {SIGKDD} Conference on
  Knowledge Discovery and Data Mining}. {ACM}, aug 2022.
\newblock \doi{10.1145/3534678.3539472}.
\newblock URL \url{https://doi.org/10.1145%2F3534678.3539472}.

\bibitem[Loynd et~al.(2020)Loynd, Fernandez, Celikyilmaz, Swaminathan, and
  Hausknecht]{WMG}
Ricky Loynd, Roland Fernandez, Asli Celikyilmaz, Adith Swaminathan, and Matthew
  Hausknecht.
\newblock Working memory graphs.
\newblock In Hal~Daumé III and Aarti Singh (eds.), \emph{Proceedings of the
  37th International Conference on Machine Learning}, volume 119 of
  \emph{Proceedings of Machine Learning Research}, pp.\  6404--6414. PMLR,
  13--18 Jul 2020.
\newblock URL \url{https://proceedings.mlr.press/v119/loynd20a.html}.

\bibitem[Lv et~al.(2021)Lv, Ding, Liu, Chen, Feng, He, Zhou, Jiang, Dong, and
  Tang]{Simple-HGN}
Qingsong Lv, Ming Ding, Qiang Liu, Yuxiang Chen, Wenzheng Feng, Siming He,
  Chang Zhou, Jianguo Jiang, Yuxiao Dong, and Jie Tang.
\newblock Are we really making much progress? revisiting, benchmarking and
  refining heterogeneous graph neural networks.
\newblock In \emph{Proceedings of the 27th ACM SIGKDD Conference on Knowledge
  Discovery \& Data Mining}, KDD '21, pp.\  1150–1160, New York, NY, USA,
  2021. Association for Computing Machinery.
\newblock ISBN 9781450383325.
\newblock \doi{10.1145/3447548.3467350}.
\newblock URL \url{https://doi.org/10.1145/3447548.3467350}.

\bibitem[McCallum et~al.(2000)McCallum, Nigam, Rennie, and Seymore]{Cora}
Andrew~Kachites McCallum, Kamal Nigam, Jason Rennie, and Kristie Seymore.
\newblock Automating the construction of internet portals with machine
  learning.
\newblock \emph{Information Retrieval}, 3:\penalty0 127--163, 2000.

\bibitem[Parisotto et~al.(2020)Parisotto, Song, Rae, Pascanu, Gulcehre,
  Jayakumar, Jaderberg, Kaufman, Clark, Noury, Botvinick, Heess, and
  Hadsell]{GTrXL}
Emilio Parisotto, Francis Song, Jack Rae, Razvan Pascanu, Caglar Gulcehre,
  Siddhant Jayakumar, Max Jaderberg, Rapha{\"e}l~Lopez Kaufman, Aidan Clark,
  Seb Noury, Matthew Botvinick, Nicolas Heess, and Raia Hadsell.
\newblock Stabilizing transformers for reinforcement learning.
\newblock In Hal~Daumé III and Aarti Singh (eds.), \emph{Proceedings of the
  37th International Conference on Machine Learning}, volume 119 of
  \emph{Proceedings of Machine Learning Research}, pp.\  7487--7498. PMLR,
  13--18 Jul 2020.
\newblock URL \url{https://proceedings.mlr.press/v119/parisotto20a.html}.

\bibitem[Rampasek et~al.(2022)Rampasek, Galkin, Dwivedi, Luu, Wolf, and
  Beaini]{GPS}
Ladislav Rampasek, Mikhail Galkin, Vijay~Prakash Dwivedi, Anh~Tuan Luu, Guy
  Wolf, and Dominique Beaini.
\newblock Recipe for a general, powerful, scalable graph transformer.
\newblock In Alice~H. Oh, Alekh Agarwal, Danielle Belgrave, and Kyunghyun Cho
  (eds.), \emph{Advances in Neural Information Processing Systems}, 2022.
\newblock URL \url{https://openreview.net/forum?id=lMMaNf6oxKM}.

\bibitem[Shaw et~al.(2018)Shaw, Uszkoreit, and Vaswani]{Shaw}
Peter Shaw, Jakob Uszkoreit, and Ashish Vaswani.
\newblock Self-attention with relative position representations.
\newblock In \emph{Proceedings of the 2018 Conference of the North {A}merican
  Chapter of the Association for Computational Linguistics: Human Language
  Technologies, Volume 2 (Short Papers)}, pp.\  464--468, New Orleans,
  Louisiana, June 2018. Association for Computational Linguistics.
\newblock \doi{10.18653/v1/N18-2074}.
\newblock URL \url{https://aclanthology.org/N18-2074}.

\bibitem[Tang et~al.(2020)Tang, Huang, Gu, Lu, and Su]{IterGNN}
Hao Tang, Zhiao Huang, Jiayuan Gu, Bao-Liang Lu, and Hao Su.
\newblock Towards scale-invariant graph-related problem solving by iterative
  homogeneous graph neural networks.
\newblock In \emph{Proceedings of the 34th International Conference on Neural
  Information Processing Systems}, NIPS'20, Red Hook, NY, USA, 2020. Curran
  Associates Inc.
\newblock ISBN 9781713829546.

\bibitem[Trask et~al.(2018)Trask, Hill, Reed, Rae, Dyer, and Blunsom]{NALU}
Andrew Trask, Felix Hill, Scott~E Reed, Jack Rae, Chris Dyer, and Phil Blunsom.
\newblock Neural arithmetic logic units.
\newblock In S.~Bengio, H.~Wallach, H.~Larochelle, K.~Grauman, N.~Cesa-Bianchi,
  and R.~Garnett (eds.), \emph{Advances in Neural Information Processing
  Systems}, volume~31. Curran Associates, Inc., 2018.
\newblock URL
  \url{https://proceedings.neurips.cc/paper/2018/file/0e64a7b00c83e3d22ce6b3acf2c582b6-Paper.pdf}.

\bibitem[Vaswani et~al.(2017)Vaswani, Shazeer, Parmar, Uszkoreit, Jones, Gomez,
  Kaiser, and Polosukhin]{VaswaniTransformer}
Ashish Vaswani, Noam Shazeer, Niki Parmar, Jakob Uszkoreit, Llion Jones,
  Aidan~N Gomez, \L~ukasz Kaiser, and Illia Polosukhin.
\newblock Attention is all you need.
\newblock In I.~Guyon, U.~Von Luxburg, S.~Bengio, H.~Wallach, R.~Fergus,
  S.~Vishwanathan, and R.~Garnett (eds.), \emph{Advances in Neural Information
  Processing Systems}, volume~30. Curran Associates, Inc., 2017.
\newblock URL
  \url{https://proceedings.neurips.cc/paper/2017/file/3f5ee243547dee91fbd053c1c4a845aa-Paper.pdf}.

\bibitem[Veli{\v{c}}kovi{\'c} et~al.(2022)Veli{\v{c}}kovi{\'c}, Badia, Budden,
  Pascanu, Banino, Dashevskiy, Hadsell, and Blundell]{CLRS}
Petar Veli{\v{c}}kovi{\'c}, Adri{\`a}~Puigdom{\`e}nech Badia, David Budden,
  Razvan Pascanu, Andrea Banino, Misha Dashevskiy, Raia Hadsell, and Charles
  Blundell.
\newblock The {CLRS} algorithmic reasoning benchmark.
\newblock In Kamalika Chaudhuri, Stefanie Jegelka, Le~Song, Csaba Szepesvari,
  Gang Niu, and Sivan Sabato (eds.), \emph{Proceedings of the 39th
  International Conference on Machine Learning}, volume 162 of
  \emph{Proceedings of Machine Learning Research}, pp.\  22084--22102. PMLR,
  17--23 Jul 2022.
\newblock URL \url{https://proceedings.mlr.press/v162/velickovic22a.html}.

\bibitem[Veli\v{c}kovi\'{c} et~al.(2020)Veli\v{c}kovi\'{c}, Buesing, Overlan,
  Pascanu, Vinyals, and Blundell]{PGN}
Petar Veli\v{c}kovi\'{c}, Lars Buesing, Matthew Overlan, Razvan Pascanu, Oriol
  Vinyals, and Charles Blundell.
\newblock Pointer graph networks.
\newblock In H.~Larochelle, M.~Ranzato, R.~Hadsell, M.F. Balcan, and H.~Lin
  (eds.), \emph{Advances in Neural Information Processing Systems}, volume~33,
  pp.\  2232--2244. Curran Associates, Inc., 2020.
\newblock URL
  \url{https://proceedings.neurips.cc/paper/2020/file/176bf6219855a6eb1f3a30903e34b6fb-Paper.pdf}.

\bibitem[Veličković et~al.(2018)Veličković, Cucurull, Casanova, Romero,
  Liò, and Bengio]{GAT}
Petar Veličković, Guillem Cucurull, Arantxa Casanova, Adriana Romero, Pietro
  Liò, and Yoshua Bengio.
\newblock Graph attention networks.
\newblock In \emph{International Conference on Learning Representations}, 2018.
\newblock URL \url{https://openreview.net/forum?id=rJXMpikCZ}.

\bibitem[Veličković et~al.(2020)Veličković, Ying, Padovano, Hadsell, and
  Blundell]{NeuralExecution}
Petar Veličković, Rex Ying, Matilde Padovano, Raia Hadsell, and Charles
  Blundell.
\newblock Neural execution of graph algorithms.
\newblock In \emph{International Conference on Learning Representations}, 2020.
\newblock URL \url{https://openreview.net/forum?id=SkgKO0EtvS}.

\bibitem[Wu et~al.(2022)Wu, Zhao, Li, Wipf, and Yan]{Nodeformer}
Qitian Wu, Wentao Zhao, Zenan Li, David Wipf, and Junchi Yan.
\newblock Nodeformer: A scalable graph structure learning transformer for node
  classification.
\newblock In Alice~H. Oh, Alekh Agarwal, Danielle Belgrave, and Kyunghyun Cho
  (eds.), \emph{Advances in Neural Information Processing Systems}, 2022.
\newblock URL \url{https://openreview.net/forum?id=sMezXGG5So}.

\bibitem[Wu et~al.(2021)Wu, Jain, Wright, Mirhoseini, Gonzalez, and
  Stoica]{GraphTrans}
Zhanghao Wu, Paras Jain, Matthew Wright, Azalia Mirhoseini, Joseph~E Gonzalez,
  and Ion Stoica.
\newblock Representing long-range context for graph neural networks with global
  attention.
\newblock In M.~Ranzato, A.~Beygelzimer, Y.~Dauphin, P.S. Liang, and J.~Wortman
  Vaughan (eds.), \emph{Advances in Neural Information Processing Systems},
  volume~34, pp.\  13266--13279. Curran Associates, Inc., 2021.
\newblock URL
  \url{https://proceedings.neurips.cc/paper/2021/file/6e67691b60ed3e4a55935261314dd534-Paper.pdf}.

\bibitem[Wu et~al.(2018)Wu, Ramsundar, Feinberg, Gomes, Geniesse, Pappu,
  Leswing, and Pande]{MoleculeNet}
Zhenqin Wu, Bharath Ramsundar, Evan N. Feinberg, Joseph Gomes, Caleb Geniesse,
  Aneesh~S. Pappu, Karl Leswing, and Vijay Pande.
\newblock Moleculenet: a benchmark for molecular machine learning.
\newblock \emph{Chem. Sci.}, 9:\penalty0 513--530, 2018.
\newblock \doi{10.1039/C7SC02664A}.
\newblock URL \url{http://dx.doi.org/10.1039/C7SC02664A}.

\bibitem[Xu et~al.(2020)Xu, Li, Zhang, Du, ichi Kawarabayashi, and
  Jegelka]{GNNAlignment}
Keyulu Xu, Jingling Li, Mozhi Zhang, Simon~S. Du, Ken ichi Kawarabayashi, and
  Stefanie Jegelka.
\newblock What can neural networks reason about?
\newblock In \emph{International Conference on Learning Representations}, 2020.
\newblock URL \url{https://openreview.net/forum?id=rJxbJeHFPS}.

\bibitem[Yadati(2020)]{G-MPNN}
Naganand Yadati.
\newblock Neural message passing for multi-relational ordered and recursive
  hypergraphs.
\newblock In H.~Larochelle, M.~Ranzato, R.~Hadsell, M.F. Balcan, and H.~Lin
  (eds.), \emph{Advances in Neural Information Processing Systems}, volume~33,
  pp.\  3275--3289. Curran Associates, Inc., 2020.
\newblock URL
  \url{https://proceedings.neurips.cc/paper/2020/file/217eedd1ba8c592db97d0dbe54c7adfc-Paper.pdf}.

\bibitem[Yang et~al.(2022)Yang, Liu, and Wang]{ReFormer}
Xuewen Yang, Yingru Liu, and Xin Wang.
\newblock Reformer: The relational transformer for image captioning.
\newblock In \emph{Proceedings of the 30th ACM International Conference on
  Multimedia}, MM '22, pp.\  5398–5406, New York, NY, USA, 2022. Association
  for Computing Machinery.
\newblock ISBN 9781450392037.
\newblock \doi{10.1145/3503161.3548409}.
\newblock URL \url{https://doi.org/10.1145/3503161.3548409}.

\bibitem[Ying et~al.(2021)Ying, Cai, Luo, Zheng, Ke, He, Shen, and
  Liu]{Graphormer}
Chengxuan Ying, Tianle Cai, Shengjie Luo, Shuxin Zheng, Guolin Ke, Di~He,
  Yanming Shen, and Tie-Yan Liu.
\newblock Do transformers really perform badly for graph representation?
\newblock In M.~Ranzato, A.~Beygelzimer, Y.~Dauphin, P.S. Liang, and J.~Wortman
  Vaughan (eds.), \emph{Advances in Neural Information Processing Systems},
  volume~34, pp.\  28877--28888. Curran Associates, Inc., 2021.
\newblock URL
  \url{https://proceedings.neurips.cc/paper/2021/file/f1c1592588411002af340cbaedd6fc33-Paper.pdf}.

\bibitem[Zaheer et~al.(2017)Zaheer, Kottur, Ravanbakhsh, Poczos, Salakhutdinov,
  and Smola]{DeepSets}
Manzil Zaheer, Satwik Kottur, Siamak Ravanbakhsh, Barnabas Poczos, Russ~R
  Salakhutdinov, and Alexander~J Smola.
\newblock Deep sets.
\newblock In I.~Guyon, U.~Von Luxburg, S.~Bengio, H.~Wallach, R.~Fergus,
  S.~Vishwanathan, and R.~Garnett (eds.), \emph{Advances in Neural Information
  Processing Systems}, volume~30. Curran Associates, Inc., 2017.
\newblock URL
  \url{https://proceedings.neurips.cc/paper/2017/file/f22e4747da1aa27e363d86d40ff442fe-Paper.pdf}.

\bibitem[Zaremba \& Sutskever(2014)Zaremba and Sutskever]{RNN-on-programs}
Wojciech Zaremba and Ilya Sutskever.
\newblock Learning to execute, 2014.
\newblock URL \url{https://arxiv.org/abs/1410.4615}.

\bibitem[Zhang et~al.(2020{\natexlab{a}})Zhang, Zhang, Xia, and
  Sun]{Graph-BERT}
Jiawei Zhang, Haopeng Zhang, Congying Xia, and Li~Sun.
\newblock Graph-bert: Only attention is needed for learning graph
  representations, 2020{\natexlab{a}}.
\newblock URL \url{https://arxiv.org/abs/2001.05140}.

\bibitem[Zhang et~al.(2020{\natexlab{b}})Zhang, Wang, Li, Zhu, Shen, Li, Lu,
  Shah, and Bennamoun]{NCI1}
Liang Zhang, Xudong Wang, Hongsheng Li, Guangming Zhu, Peiyi Shen, Ping Li,
  Xiaoyuan Lu, Syed Afaq~Ali Shah, and Mohammed Bennamoun.
\newblock Structure-feature based graph self-adaptive pooling.
\newblock In \emph{Proceedings of The Web Conference 2020}. {ACM}, apr
  2020{\natexlab{b}}.
\newblock \doi{10.1145/3366423.3380083}.
\newblock URL \url{https://doi.org/10.1145%2F3366423.3380083}.

\bibitem[Zhang et~al.(2020{\natexlab{c}})Zhang, Liu, and Xie]{MXMNet}
Shuo Zhang, Yang Liu, and Lei Xie.
\newblock Molecular mechanics-driven graph neural network with multiplex graph
  for molecular structures.
\newblock In \emph{NeurIPS-W}, 2020{\natexlab{c}}.

\end{thebibliography}
\bibliographystyle{iclr2023_conference}

\newpage

\appendix

\section{Transformers}
\label{section:appendix_transformer}

We describe the transformer architecture introduced by \citet{VaswaniTransformer}. This description also applies to most transformer variants proposed over the years. Layer superscripts are employed to distinguish input vectors from output vectors, and are often omitted for vectors inside the same layer.

Although the transformer is a set-to-set model, it can be described using our graph-to-graph formalism as limited to computation over \textit{nodes} only.
Each transformer layer is a function passing updated node vectors to the next layer. A single transformer layer can therefore be expressed as a modified version of \eqref{node update}:
\begin{equation}
    \mathbf{n}^{l + 1}_{i} = \phi_{n} \left(\mathbf{n}^{l}_{i}, \bigoplus \limits_{j \in \mathcal{L}_{i}}
    a \left(\mathbf{n}^{l}_{i}, \mathbf{n}^{l}_{j}\right) \psi^{m} \left(\mathbf{n}^{l}_{j}\right) \right)
    \label{eq:standardatt}
\end{equation}

where $a \left(\mathbf{n}^{l}_{i}, \mathbf{n}^{l}_{j}\right)$ computes the attentional coefficient $\alpha^l_{ij}$ applied by node $i$ to the value vector $\mathbf{v}^l_j$, which is computed by $\psi^{m} \left(\mathbf{n}^{l}_{j}\right)$, a linear transformation:
\begin{equation}
    \alpha^l_{ij} = a \left(\mathbf{n}^{l}_{i}, \mathbf{n}^{l}_{j}\right) \qquad \mathbf{v}^l_j = \psi^{m} \left(\mathbf{n}^{l}_{j}\right) = \mathbf{n}^l_j {W^V} \label{QKV}
\end{equation}

The attentional coefficients applied by node $i$ to the set of all nodes $j$ is a probability distribution $\mathbf{a}_i$ computed by the softmax function over a set of vector dot products:
\begin{equation}
  \mathbf{a}_i = \text{softmax}_j \left(\frac{\mathbf{q}_i \mathbf{k}_j^T}{\sqrt{d_{n}}}\right) \label{dotprod}
\end{equation}

The \textit{QKV} vectors introduced above are linear transformations of node vectors:
\begin{equation}
  \mathbf{q}_i = \mathbf{n}_i {W^Q} \qquad \mathbf{k}_j = \mathbf{n}_j {W^K} \qquad \mathbf{v}_j = \mathbf{n}_j {W^V} \label{QKV2}
\end{equation}

where $\mathbf{W}^Q \in \R^{d_n {\times} d_n}$,
$\mathbf{W}^K \in \R^{d_n {\times} d_n}$,
$\mathbf{W}^V \in \R^{d_n {\times} d_n}$, and $d_n$ is the node vector size. 
These $\mathbf{W}$ matrices (like all other trainable parameters in $\phi_{n}$) are not shared between transformer layers.

The aggregation function $\bigoplus$ sums the incoming messages from all nodes $\mathcal{L}_{i}$ in the completely connected graph:
\begin{equation}
    \mathbf{m}^l_{i} = \bigoplus \limits_{j \in \mathcal{L}_{i}}
    a \left(\mathbf{n}^{l}_{i}, \mathbf{n}^{l}_{j}\right) \psi^{m} \left(\mathbf{n}^{l}_{j}\right)
    = \sum \limits_{j} \alpha^l_{ij} \mathbf{v}^l_j
\end{equation}

This aggregated message $\mathbf{m}^l_i$ is then passed to the local update function $\phi_{n}$ (shared by all nodes), which is the following stack of linear layers, skip connections, layer normalization and a ReLU activation function:
\begin{align}
    \mathbf{u}^l_i &= \text{LayerNorm} \left(\mathbf{m}^l_i \mathbf{W}_1 + \mathbf{n}^{l}_{i}\right) \\
    \mathbf{n}^{l+1}_{i} &= \text{LayerNorm} \left(\text{ReLU} (\mathbf{u}^l_i \mathbf{W}_2) \mathbf{W}_3 + \mathbf{u}^l_i\right)
    \label{node_update}
\end{align}

where $\mathbf{W}_1 \in \R^{d_n {\times} d_n}$,
$\mathbf{W}_2 \in \R^{d_n {\times} d_{nh}}$,
$\mathbf{W}_3 \in \R^{d_{nh} {\times} d_n}$, and $d_{nh}$ is the hidden layer size of the feed-forward network.

This overview of the transformer architecture has focused on the fully connected case of self-attention. For brevity we have omitted the details of multi-head attention, bias vectors, and the stacking of vectors into matrices for maximal GPU utilization. 

\section{Sokoban Experiments}
\label{section:appendix_sokoban}
In the experiments described in the main text, the model received edge feature vectors as inputs. 
The question we pose here is whether RT's latent edge vectors can improve reasoning ability even on tasks with graph structure that is \textit{hidden}, rather than passed to the model in the form of edge vectors.
We use the Sokoban \citep{Sokoban} reinforcement learning task to investigate.
In Sokoban, the agent must push four yellow boxes onto the red targets within 120 time steps.
Humans solving these puzzles tend to plan out which boxes will go onto which targets.
Assuming that a successful agent will learn a similar strategy, representing each box-to-target pair as a directed relation, we hypothesize that RT is more capable than a standard transformer at reasoning over such pairwise relations.
For evaluation, we use RT in place of the standard transformer originally used by the Working Memory Graph (WMG) RL agent \citep{WMG}, then train both modified and unmodified agents on the Sokoban task for 10 million time steps. 

\textbf{Main Results} We find that using RT reduces the agent's final error rate by a relative \textbf{15\%}, from 34\% to 29\% of the puzzles. This improvement is much larger than the confidence intervals (0.3\% and 0.7\% standard error, respectively). Our results support the hypothesis that RT can learn even hidden graph structure.
One alternative explanation would be that the extra trainable parameters added by RT simply improved the expressivity of WMG.
But this seems unlikely since hyperparameter tuning of the unmodified WMG agent \citep{WMG} converged to an intermediate model size, rather than a larger model for more expressivity.

\section{Hyperparameters}
\label{section:appendix_hps}

To tune the hyperparameters of RT and the CLRS-30 baseline GNNs, we used Distributed Grid Descent (DGD) \citep{WMG}, a self-guided form of random search.
Each search was terminated after model performance converged to a stable value. Then 20 additional runs were executed, using the winning hyperparameter configuration, to obtain results free from selection bias.
All tuning runs used the CLRS-30 protocol described in \ref{subsubsection:dataset_enhance}, except for the following details:

\begin{enumerate}
    \item To prevent tuning on the canonical datasets or any fixed datasets at all, the dataset generation seeds were randomized at the start of each run.
    \item To reduce variance, the minimum evaluation dataset size was raised from 32 to 100.
    \item To mitigate the computational costs, all models were tuned on only the 8 core algorithms, and each training run was shortened to 32,000 examples.
\end{enumerate}

Very similar procedures were used to tune RT hyperparameters for the other (non-CLRS-30) experiments. For all experiments, all untuned hyperparameter values were chosen to match the settings of the corresponding baseline models.

\subsection{CLRS-30 experiments}

Table \ref{clrs_hp_table} lists the tuned hyperparameter values for CLRS-30 experiments, and
Table \ref{all_hp_values_table} reports the sets of values considered in those searches.
The runtime sizes of the corresponding models are found in Table \ref{clrs_sizes}, along with their training speeds in Table \ref{clrs_speeds}.

\begin{table*}[htb]
\small
\caption{
Tuned hyperparameter values for CLRS-30 experiments.
}
\label{clrs_hp_table}
\begin{center}
\begin{tabular}{l | r r r r r r r}
	& \textbf{Deep Sets}	&	\textbf{GAT-v1}	&	\textbf{GAT-v2}	&	\textbf{MPNN}	&	\textbf{PGN-u}	&	\textbf{PGN-m}	&	\textbf{RT}	\\
\hline
batch\_size     & 4 & 4 & 4 & 4 & 4 & 4 & 4 \\
num\_layers	     & 2	&	2	&	2	&	1	&	1	&	3	&	3	\\
learning\_rate	 & 2.5e-4	&	6.3e-4	&	2.5e-4	&	1.6e-3	&	1.6e-3	&	1e-3	&	2.5e-4	\\
$d_n=d_e=d_g$	 & 512	&	-	&	-	&	512	&	180	&	512	&	-	\\
nb\_heads        & -	&	10	&	12	&	-	&	-	&	-	&	12	\\
head\_size       & -	&	64	&	64	&	-	&	-	&	-	&	16	\\
$d_{nh}$             & -	&	-	&	-	&	-	&	-	&	-	&	32	\\
$d_{eh1}$	         & -	&	-	&	-	&	-	&	-	&	-	&	16	\\
$d_{eh2}$	         & -	&	-	&	-	&	-	&	-	&	-	&	8	\\
ptr\_from\_edges & -	&	-	&	-	&	-	&	-	&	-	&	true	\\
graph\_vec       & -	&	-	&	-	&	-	&	-	&	-	&	cat	\\
\hline
\end{tabular}
\end{center}
\end{table*}

\begin{table*}[htb]
\small
\caption{
Hyperparameter values considered for CLRS-30 experiments.
}
\label{all_hp_values_table}
\begin{center}
\begin{tabular}{l | l}
\hline
batch\_size      &  1, 2, 4, 8, 16 \\
num\_layers	     &	1, 2, 3, 4	\\
learning\_rate	 &	4e-5, 6.3e-5, 1e-4, 1.6e-4, 2.5e-4, 4e-4, 6.3e-4, 1e-3, 1.6e-3, 2.5e-3, 4e-3, 6.3e-3, 1e-2	\\
$d_n=d_e=d_g$	 &	45, 64, 90, 128, 180, 256, 360, 512	\\
nb\_heads        &  3, 4, 6, 8, 10, 12, 16 \\
head\_size       &  8, 12, 16, 24, 32, 45, 64 \\
$d_{nh}$              &  4, 6, 8, 12, 16, 24, 32, 45, 64, 90 \\
$d_{eh1}$	         &	12, 16, 24, 32, 45, 64	\\
$d_{eh2}$	         &	4, 6, 8, 12, 16, 24, 32, 45	\\
ptr\_from\_edges &	false, true	\\
graph\_vec       &  core, cat	\\
\hline
\end{tabular}
\end{center}
\end{table*}

\begin{table*}[htb]
\small
\caption{
Number of trainable parameters in each model tested on CLRS-30, including the framework's encoding and decoding layers, on the reference algorithm Bellman Ford.
}
\label{clrs_sizes}
\begin{center}
\begin{tabular}{l | r r r r r r r}
	& \textbf{Deep Sets}	&	\textbf{GAT-v1}	&	\textbf{GAT-v2}	&	\textbf{MPNN}	&	\textbf{PGN-u}	&	\textbf{PGN-m}	&	\textbf{RT}	\\
\hline
tuned	     & 8,675,332	&	11,168,732	&	17,730,076	&	6,573,060	&	816,844	&	10,777,604	&	1,103,404	\\
untuned	 & 414,468	&	418,960	&	397,957	&	414,468	&	414,468	&	414,468	&	-	\\
\hline
\end{tabular}
\end{center}
\end{table*}

\begin{table*}[h!]
\small
\caption{
Training speed in examples per second on a T4 GPU, on the reference algorithm Bellman Ford.
}
\label{clrs_speeds}
\begin{center}
\begin{tabular}{l | r r r r r r r}
	& \textbf{Deep Sets}	&	\textbf{GAT-v1}	&	\textbf{GAT-v2}	&	\textbf{MPNN}	&	\textbf{PGN-u}	&	\textbf{PGN-m}	&	\textbf{RT}	\\
\hline
tuned	     & 13.4	&	14.8	&	10.1	&	15.7	&	18.7	&	11.6	&	8.7	\\
untuned	 & 128.9	&	140.7	&	134.8	&	128.8	&	128.6	&	128.6	&	-	\\
\hline
\end{tabular}
\end{center}
\end{table*}

\newpage

\subsection{Lobster graph hyperparameters}

Table \ref{lobster_hp_table} lists the tuned hyperparameter values for RT on the shortest-path task over lobster graphs, along with the sets of values considered in that search.

\begin{table*}[htb]
\small
\caption{
Tuned hyperparameter values for RT on the lobster graph experiment.
}
\label{lobster_hp_table}
\begin{center}
\begin{tabular}{l | r l }
	& \textbf{Tuned values}	&	\textbf{Values considered}	\\
\hline
learning\_rate	 & 6.3e-5	&	4e-5, 6.3e-5, 1e-4, 1.6e-4, 2.5e-4, 4e-4	\\
output\_layer\_size  & 180 & 64, 90, 128, 180, 256, 360, 512 \\
nb\_heads        & 8	&	3, 4, 6, 8, 10		\\
head\_size       & 28	&	12, 16, 20, 24, 28, 32		\\
$d_{nh}$             & 12	&	3, 4, 6, 8, 12, 16, 24		\\
$d_e$	 & 128	&	16, 24, 32, 45, 64, 90, 128		\\
$d_{eh1}$	         & 32	&	12, 16, 24, 32, 45, 64, 90		\\
$d_{eh2}$	         & 8	&	8, 12, 16, 24, 32, 45, 64		\\
dropout\_rate & 0.0	&	0.0, 0.01, 0.02, 0.04, 0.08		\\
grad\_clip       & 128.0	&	16.0, 32.0, 64.0, 128.0, 256.0, 512.0		\\
\hline
\end{tabular}
\end{center}
\end{table*}

\subsection{Sokoban hyperparameters}

Table \ref{sokoban_hp_table} lists the tuned hyperparameter values for RT on the Sokoban task, along with the sets of values considered in that search.

\begin{table*}[htb]
\small
\caption{
Tuned hyperparameter values for RT on Sokoban.
}
\label{sokoban_hp_table}
\begin{center}
\begin{tabular}{l | r l }
	& \textbf{Tuned values}	&	\textbf{Values considered}	\\
\hline
$d_e$	 & 128	&	8, 12, 16, 24, 32, 45, 64, 90, 128, 180, 256, 360, 512, 720, 1024		\\
$d_{eh1}$	         & 64	&	6, 8, 12, 16, 24, 32, 45, 64, 90, 128, 180, 256, 360, 512		\\
$d_{eh2}$	         & 12	&	6, 8, 12, 16, 24, 32, 45, 64, 90, 128, 180, 256, 360, 512		\\
\hline
\end{tabular}
\end{center}
\end{table*}

\section{Train/Test Protocol}
\label{subsubsection:dataset_enhance}
Except where specifically noted, our experiments follow the exact train/test protocol defined by CLRS-30.
CLRS-30 provides canonical datasets (training, validation, and test) which can also be generated from specific random seeds: 1, 2, 3.
The graphs in the training and validation datasets contain 16 nodes, while the test graphs are of size 64 to evaluate the out-of-distribution (OOD) generalization of models.
Training is performed on a random sequence of 320,000 examples drawn with replacement from the train set, where each example trajectory contains a variable number of reasoning steps.
During training, the model is evaluated on the validation set after every 320 examples.
At the end of training, the model with the highest validation score is evaluated on the test set.
The average test micro-F1 score is reported for all results.
The published CLRS-30 results are reported as averages over 3 random run seeds, but we use 20 seeds in all of our experiments.

\section{Test Results on CLRS-30}
\label{section:appendix_clrs}

Test performance of all tuned models evaluated on CLRS-30 may be found in Tables \ref{tab:clrs_meanacc} (mean test micro-F1 score) and \ref{tab:clrs_meandev} (standard deviation). On certain tasks in which baseline GNN performances improved, we also observed higher variability in results. Following \citet{CLRS}, we report the best performance between PGN-u and PGN-m on every task in the column titled PGN-c (for PGN-combination). Note therefore that PGN-c does not represent a single model.

We bold the results of the best-performing single model for each specific task. RT was the best-performing on \textbf{11 out of 30} of the tasks. Overall, MPNN was the top-scoring baseline model, winning on 8 tasks. Deep Sets won on 4 tasks. PGN-u won on 1 task, and PGN-m won on 5 tasks. Finally, GAT-v2 won on 1 task.

Table \ref{tab:clrs_classes} shows test performances of all tuned models across 8 algorithm classes. As mentioned in Section \ref{subsubsection:clrs_results}, RT performed best in \textbf{6 out of 8} of the classes, and second-best on greedy algorithms and sorting.

Table \ref{tab:clrs_catmapping} shows which algorithms belong to which classes, and which task types. Each algorithm may correspond to multiple task types if the hint and final outputs of that algorithm differ.

Table \ref{tab:clrs_size} shows test performance of the baseline GNN models trained on datasets of varying size, without hyperparameter tuning. Specifically, we compare test scores between the models trained on datasets of size 1000, and datasets of size 10000. We note that the results in the first row agree closely with those in the CLRS-30 paper (Table \ref{tab:clrs_repr}).

\section{CLRS-30 Benchmark Ablations}
\label{section:appendix_abl}

We perform our ablation studies on the 8 core CLRS-30 tasks. Table \ref{tab:clrs_pt} compares the test performance of the standard transformer to the performance of RT. Table \ref{tab:clrs_layers} shows the drop in test performance that resulted from restricting RT to one layer. The original, tuned RT had 3 layers, and is labeled RT-3. Table \ref{tab:clrs_decode} shows the marginal improvement that resulted from decoding node vectors using only edge vectors. Table \ref{tab:clrs_core} shows the effects that handling global input vectors through a core node (vs. concatenation with the input node vectors) has on RT test performance. RT with core node only won on 3 out of 7 tasks, but had the higher average test performance by 0.08\%. Finally, Table \ref{tab:clrs_edgeupdate} shows the drop in test performance that resulted from removing edge updates from RT.

\section{CLRS-30 Model Variance}
\label{section:appendix_variance}

\begin{figure}[!b]
    \centering
    \caption{Histograms describing the variance in model results on Naïve String Matcher, across runs}
    \includegraphics[width=1.0\linewidth]{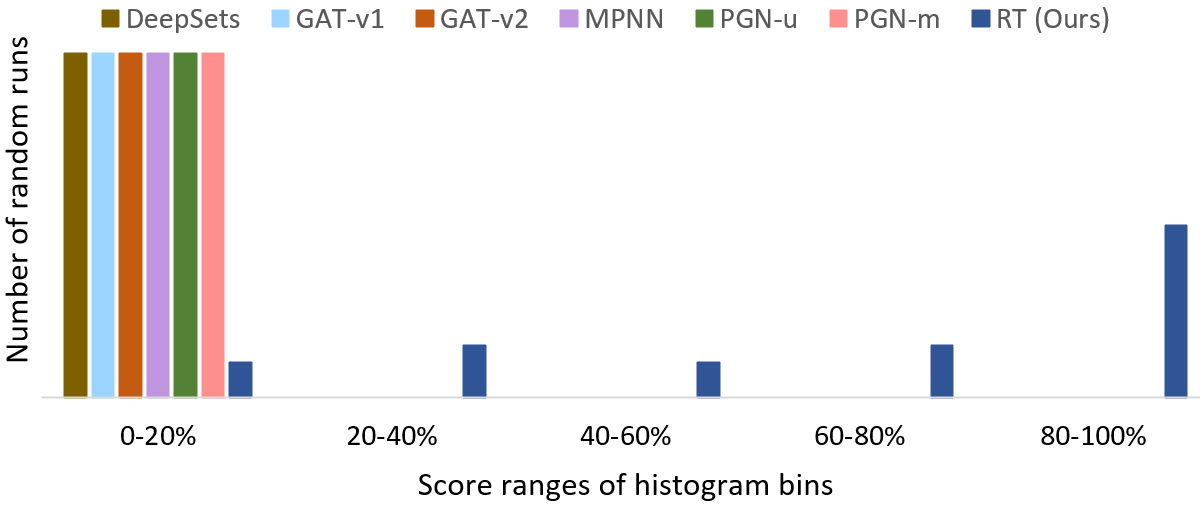}
    \label{fig:histogram_nsm}
\end{figure}

Table \ref{tab:clrs_meandev} shows that RT produces some of the largest standard deviations over random runs. To investigate these results, we first examine the Naïve String Matcher task, where RT's variance is the largest value in the table (32.3\%). Figure \ref{fig:histogram_nsm} displays each model's distribution of run scores, through histograms. We see that RT's twenty runs on this task obtain a wide range of scores, including a spike in the range from 80\% to 100\%, while none of the baseline model runs surpass a score of 20\%. This explains RT's high variance on this task compared to the other models. In summary, in a case like this higher variance is a consequence of higher performance.

The same behavior can be inferred without the histograms by examining Table \ref{tab:clrs_meandev} and Table \ref{tab:clrs_meanacc}, which show that RT's high variance on Naïve String Matcher is associated with far higher mean performance on the task than the baseline models. The same pattern is apparent for RT's next two highest-variance tasks, Topological Sort and Quickselect.

\begin{figure}[!b]
    \centering
    \caption{Histograms describing the variance in model results across all runs on the 30 CLRS tasks}
    \includegraphics[width=1.0\linewidth]{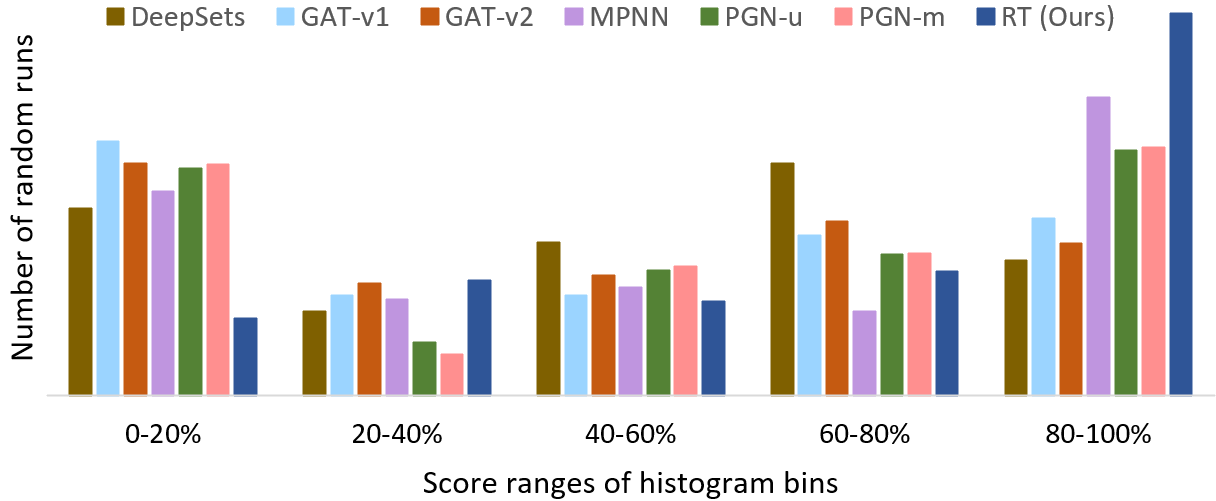}
    \label{fig:histogram_30}
\end{figure}

For a broader view of variance, Figure \ref{fig:histogram_30} displays the model score distributions over all CLRS-30 tasks. We see that RT's distribution is weighted more heavily than the other models in the $>$80\% range, and underweighted in the $<$20\% range. Meanwhile, all model distributions are spread quite widely over the five range bins. This spread is quantified numerically by the last row of Table \ref{tab:clrs_meandev}, where RT's overall standard deviation is shown to be one of the smallest.

































\newpage

\begin{table}[tb]
\caption{Algorithms and their respective classes and task types. Taking hint targets into consideration, there are 3 task types in CLRS-30: node (N), edge (E), and graph (G).
}
\label{tab:clrs_catmapping}

\small
\centering
\noindent
\begin{adjustbox}{width=1.0\linewidth, center}
\begin{tabular}{| l | c | c| c | c | c | c | c | c | c | c | c |}\hline
{\textbf{Algorithm}} & {\textbf{N}} & {\textbf{E}} & {\textbf{G}} & {\textbf{D\&C}} & {\textbf{DP}} & {\textbf{Geo}} & {\textbf{Graph}} & {\textbf{Greedy}} & {\textbf{Search}} & {\textbf{Sort}} & {\textbf{String}} \\ \hline

Activity Selector  & \cellcolor{gray} & \cellcolor{gray} & & & & & & \cellcolor{black} & & &\\

\hline

Articulation Points & \cellcolor{gray} & \cellcolor{gray} & \cellcolor{gray} & & & & \cellcolor{black} & & & &  \\ 

\hline

Bellman-Ford & \cellcolor{gray} & \cellcolor{gray} & & & & & \cellcolor{black} & & & & \\

\hline

BFS & \cellcolor{gray} & \cellcolor{gray} & & & & & \cellcolor{black} & & & & \\

\hline

Binary Search & \cellcolor{gray} & \cellcolor{gray} & & & & & & & \cellcolor{black} & & \\

\hline

Bridges & \cellcolor{gray} & \cellcolor{gray} & \cellcolor{gray} & & & & \cellcolor{black} & & & & \\

\hline

Bubble Sort & \cellcolor{gray} & \cellcolor{gray} & & & & & & & & \cellcolor{black} & \\

\hline

DAG Shortest Paths & \cellcolor{gray} & \cellcolor{gray} & \cellcolor{gray} & & & & \cellcolor{black} & & & & \\

\hline

DFS & \cellcolor{gray} & \cellcolor{gray} & \cellcolor{gray} & & & & \cellcolor{black} & & & & \\

\hline

Dijkstra & \cellcolor{gray} & \cellcolor{gray} & & & & & \cellcolor{black} & & & & \\

\hline

Find Max. Subarray & \cellcolor{gray} & \cellcolor{gray} & \cellcolor{gray} & \cellcolor{black} & & & & & & & \\

\hline

Floyd-Warshall & \cellcolor{gray} & \cellcolor{gray} & & & & & \cellcolor{black} & & & & \\

\hline

Graham Scan & \cellcolor{gray} & \cellcolor{gray} & \cellcolor{gray} & & & \cellcolor{black} & & & & &  \\

\hline

Heapsort & \cellcolor{gray} & \cellcolor{gray} & \cellcolor{gray} & & & & & & & \cellcolor{black} & \\

\hline

Insertion Sort & \cellcolor{gray} & \cellcolor{gray} & & & & & & & & \cellcolor{black} & \\

\hline

Jarvis' March & \cellcolor{gray} & \cellcolor{gray} & \cellcolor{gray} & & & \cellcolor{black} & & & & & \\

\hline

KMP Matcher & \cellcolor{gray} & \cellcolor{gray} & \cellcolor{gray} & & & & & & & & \cellcolor{black} \\

\hline

LCS Length & & \cellcolor{gray} & & & \cellcolor{black} & & & & & & \\

\hline

Matrix Chain Order & & \cellcolor{gray} & & & \cellcolor{black} & & & & & & \\

\hline

Minimum & \cellcolor{gray} & \cellcolor{gray} & & & & & & & \cellcolor{black} & & \\

\hline

MST-Kruskal & \cellcolor{gray} & \cellcolor{gray} & \cellcolor{gray} & & & & \cellcolor{black} & & & & \\

\hline

MST-Prim & \cellcolor{gray} & \cellcolor{gray} & & & & & \cellcolor{black} & & & & \\

\hline

Na\"{i}ve String Match & \cellcolor{gray} & \cellcolor{gray} & & & & & & & & & \cellcolor{black} \\

\hline

Optimal BST & & \cellcolor{gray} & & & \cellcolor{black} & & & & & & \\

\hline

Quickselect & \cellcolor{gray} & \cellcolor{gray} & \cellcolor{gray} & & & & & & \cellcolor{black} & & \\

\hline

Quicksort & \cellcolor{gray} & \cellcolor{gray} & & & & & & & & \cellcolor{black} & \\

\hline

Segments Intersect & \cellcolor{gray} & & \cellcolor{gray} & & & \cellcolor{black} & & & & & \\

\hline

SCC & \cellcolor{gray} & \cellcolor{gray} & \cellcolor{gray} & & & & \cellcolor{black} & & & & \\

\hline

Task Scheduling & \cellcolor{gray} & \cellcolor{gray} & \cellcolor{gray} & & & & & \cellcolor{black} & & & \\

\hline

Topological Sort & \cellcolor{gray} & \cellcolor{gray} & & & & & \cellcolor{black} & & & & \\

\hline

Total & 27 & 29 & 14 & 1 & 3 & 3 & 12 & 2 & 3 & 4 & 2 \\

\hline

\end{tabular}%
\end{adjustbox}
\end{table}

\begin{table*}[b]
\renewcommand{\arraystretch}{1.0}
\caption{Average test scores of untuned baseline models trained on either the original small datasets, versus trained on the expanded training datasets, for the 8 core tasks.
}
\label{tab:clrs_size}

\small

\begin{center}
\begin{tabular}{lrrrrrrr}\hline
{\textbf{Training Set Size}} & {\textbf{Deep Sets}} & {\textbf{GAT-v1}} & {\textbf{GAT-v2}} & {\textbf{MPNN}} & {\textbf{PGN-u}} & {\textbf{PGN-m}} & {\textbf{PGN-c}} \\ \hline

1000 & 45.95\% & 42.24\% & 42.53\% & 54.66\% & 51.07\% & {56.18\%} & 61.29\% \\

10000 & 46.70\% & 42.97\% & 45.37\% & 56.93\% & 56.77\% & {58.20\%} & 67.11\% \\

\hline
\end{tabular}
\end{center}
\end{table*}

\begin{table*}[t]
\renewcommand{\arraystretch}{1.0}
\caption{Test score improvements from hyperparameter tuning on the 8 core tasks. The drop in score for PGN-m was likely the result of tuning hyperparameters (for all models) on shorter training runs than was used for evaluation.
}
\label{tab:clrs_tuning_2}

\small

\begin{center}
\begin{tabular}{lrrrrrrr}\hline
{\textbf{Algorithm}} & {\textbf{Deep Sets}} & {\textbf{GAT-v1}} & {\textbf{GAT-v2}} & {\textbf{MPNN}} & {\textbf{PGN-u}} & {\textbf{PGN-m}} & {\textbf{PGN-c}} \\ \hline

No Tuning & 46.70\% & 42.97\% & 45.37\% & 56.93\% & 56.77\% & 58.20\% & 67.11\% \\

Tuning & 51.55\% & 50.47\% & 48.97\% & 59.75\% & 59.97\% & 57.53\% & 67.86\% \\

 \hline
\end{tabular}
\end{center}
\end{table*}

\begin{table*}[b]
\renewcommand{\arraystretch}{1.0}
\caption{Reproduction of baseline model results on the 8 core algorithms.
}
\label{tab:clrs_repr}

\small

\begin{center}
\begin{tabular}{llrrr}\hline
{\textbf{Results}} & {\textbf{\# Seeds}} & {\textbf{Deep Sets}} & {\textbf{MPNN}} & {\textbf{PGN-c}}  \\ \hline

Published \citep{CLRS} & 3 & 45.50\% & 53.05\% & 61.15\%   \\

Reproduced (Ours) & 20 & 45.95\% & 54.66\% & 61.29\%  \\

\hline
\end{tabular}
\end{center}
\end{table*}

\begin{table*}[b]
\renewcommand{\arraystretch}{1.0}
\caption{Test score improvements from augmenting the CLRS-30 baseline models.
}
\label{tab:clrs_improve_base}

\small

\begin{center}
\begin{tabular}{llrrr}\hline
{\textbf{Results}} & {\textbf{\# Seeds}} & {\textbf{Deep Sets}} & {\textbf{MPNN}} & {\textbf{PGN-c}}\\ \hline

CLRS-30 Published Results & 3 & 42.72\% & 44.99\% & 50.84\% \\

Our Results & 20 & 48.60\% & 49.23\% & 52.82\% \\

\hline
\end{tabular}
\end{center}
\end{table*}

\begin{table*}[tb]
\renewcommand{\arraystretch}{1.0}
\caption{Standard transformer ablation (with re-tuned hyperparameters) evaluated on the 8 core algorithms.
}
\label{tab:clrs_pt}

\small

\begin{center}
\begin{tabular}{lrr}\hline
{\textbf{Algorithm}} & {\textbf{Transformer}} & \textbf{RT (Ours)} \\ \hline

Activity Selector & 68.01\% $\pm$ 6.0 & {\textbf{87.72\%}} $\pm$ 2.7 \\

Bellman-Ford & 39.79\% $\pm$ 2.4 & {\textbf{94.24\%}} $\pm$ 1.5 \\

Binary Search & 2.84\% $\pm$ 0.6 & {\textbf{81.48\%}} $\pm$ 6.7 \\

Find Max. Subarray & 26.60\% $\pm$ 6.0 & {\textbf{66.52\%}} $\pm$ 3.7 \\

Graham Scan & 58.84\% $\pm$ 6.9 & {\textbf{74.15\%}} $\pm$ 7.4 \\

Insertion Sort & 60.57\% $\pm$ 15.7 & {\textbf{89.43\%}} $\pm$ 9.0 \\

Matrix Chain Order & 80.54\% $\pm$ 6.8 & {\textbf{91.89\%}} $\pm$ 1.2 \\

Na\"{i}ve String Match & 1.51\% $\pm$ 1.7 & {\textbf{65.01\%}} $\pm$ 32. \\

\hline

Average & 42.34\% & {\textbf{81.30\%}} \\

 \hline
\end{tabular}
\end{center}
\end{table*}

\begin{table*}[tb]
\renewcommand{\arraystretch}{1.0}
\caption{Single-layer RT ablation (with re-tuned hyperparameters) evaluated on the 8 core algorithms.
}
\label{tab:clrs_layers}

\small

\begin{center}
\begin{tabular}{lrr}\hline
{\textbf{Algorithm}} & {\textbf{RT-1}} & {\textbf{RT-3 (Ours)}} \\ \hline

Activity Selector & 83.41\% $\pm$ 5.8 & {\textbf{87.72\%}} $\pm$ 2.7 \\

Bellman-Ford & 93.52\% $\pm$ 1.7 & {\textbf{94.24\%}} $\pm$ 1.5 \\

Binary Search & 61.61\% $\pm$ 15.1 & {\textbf{81.48\%}} $\pm$ 6.7 \\

Find Max. Subarray & 65.07\% $\pm$ 4.0 & {\textbf{66.52\%}} $\pm$ 3.7 \\

Graham Scan & \textbf{74.80\%} $\pm$ 7.9 & {74.15\%} $\pm$ 7.4 \\

Insertion Sort & 77.54\% $\pm$ 10.3 & {\textbf{89.43\%}} $\pm$ 9.0 \\

Matrix Chain Order & 91.05\% $\pm$ 1.0 & {\textbf{91.89\%}} $\pm$ 1.2 \\

Na\"{i}ve String Match & 16.50\% $\pm$ 23.2 & {\textbf{65.01\%}} $\pm$ 32. \\

\hline

Average & 70.44\% & {\textbf{81.30\%}} \\

 \hline
\end{tabular}
\end{center}
\end{table*}

\begin{table*}[tb]
\renewcommand{\arraystretch}{1.0}
\caption{Ablation of the node pointer decoding procedure evaluated on the 8 core algorithms.
}
\label{tab:clrs_decode}

\small

\begin{center}
\begin{tabular}{lrr}\hline
{\textbf{Algorithm}} & {\textbf{RT with original decoding}} & {\textbf{RT with edge-only decoding (Ours)}} \\ \hline

Activity Selector & {\textbf{88.09\%}} $\pm$ 5.0 & 87.72\% $\pm$ 2.7 \\

Bellman-Ford & {\textbf{94.95\%}} $\pm$ 1.1 & 94.24\% $\pm$ 1.5 \\

Binary Search & 80.06\% $\pm$ 6.6 & {\textbf{81.48\%}} $\pm$ 6.7 \\

Find Max. Subarray & {\textbf{67.44\%}} $\pm$ 3.7 & 66.52\% $\pm$ 3.7 \\

Graham Scan & {\textbf{75.71\%}} $\pm$ 10.3 & 74.15\% $\pm$ 7.4 \\

Insertion Sort & 71.32\% $\pm$ 11.0 & {\textbf{89.43\%}} $\pm$ 9.0 \\

Matrix Chain Order & {\textbf{92.12\%}} $\pm$ 0.9 & 91.89\% $\pm$ 1.2 \\

Na\"{i}ve String Match & {\textbf{78.43\%}} $\pm$ 22.5 & 65.01\% $\pm$ 32. \\

\hline

Average & 81.00\% & {\textbf{81.30\%}} \\

 \hline
\end{tabular}
\end{center}
\end{table*}

\begin{table*}[tb]
\renewcommand{\arraystretch}{1.0}
\caption{Ablation of RT core node (vs. concatenation of the global vector) evaluated on 7 representative algorithms that use global input features or hints.
}
\label{tab:clrs_core}

\small

\begin{center}
\begin{tabular}{lrr}\hline
{\textbf{Algorithm}} & {\textbf{RT with core node}} & {\textbf{RT without core node (Ours)}} \\ \hline

Binary Search & 75.40\% $\pm$ 11.0 & {\textbf{81.48\%}} $\pm$ 6.7 \\

Find Max. Subarray & {\textbf{66.96\%}} $\pm$ 4.5 & 66.52\% $\pm$ 3.7 \\

Graham Scan & 71.83\% $\pm$ 10.2 & {\textbf{74.15\%}} $\pm$ 7.4 \\

Heapsort & 30.67\% $\pm$ 18.9 & {\textbf{32.96\%}} $\pm$ 14.8 \\

KMP Matcher & 0.02\% $\pm$ 0.0 & {\textbf{0.03\%}} $\pm$ 0.1 \\

MST-Kruskal & {\textbf{75.59\%}} $\pm$ 5.8 & 64.91\% $\pm$ 11.8 \\

Task Scheduling & {\textbf{83.09\%}} $\pm$ 1.8 & 82.93\% $\pm$ 1.8 \\

\hline

Average & {\textbf{57.65\%}} & 57.57\% \\

\hline
\end{tabular}
\end{center}
\end{table*}

\begin{table*}[t]
\renewcommand{\arraystretch}{1.0}
\caption{Ablation of RT's edge update procedure evaluated on the 8 core algorithms
}
\label{tab:clrs_edgeupdate}

\small

\begin{center}
\begin{tabular}{lrr}\hline
{\textbf{Algorithm}} & {\textbf{RT without edge updates}} & {\textbf{RT with edge updates (Ours)}} \\ \hline

Activity Selector & 82.13\% $\pm$ 7.0 & {\textbf{87.72\%}} $\pm$ 2.7 \\

Bellman-Ford & 91.03\% $\pm$ 2.3 & {\textbf{94.24\%}} $\pm$ 1.5 \\

Binary Search & 60.29\% $\pm$ 12.5 & {\textbf{81.48\%}} $\pm$ 6.7 \\

Find Max. Subarray & 19.09\% $\pm$ 1.6 & {\textbf{66.52\%}} $\pm$ 3.7 \\

Graham Scan & 69.19\% $\pm$ 7.7 & {\textbf{74.15\%}} $\pm$ 7.4 \\

Insertion Sort & 22.30\% $\pm$ 8.5 & {\textbf{89.43\%}} $\pm$ 9.0 \\

Matrix Chain Order & 85.68\% $\pm$ 1.5 & {\textbf{91.89\%}} $\pm$ 1.2 \\

Na\"{i}ve String Match & 2.22\% $\pm$ 0.8 & {\textbf{65.01\%}} $\pm$ 32. \\

\hline

Average & 53.99\% & {\textbf{81.30\%}} \\

 \hline
\end{tabular}
\end{center}
\end{table*}

\begin{table*}[tb]
\renewcommand{\arraystretch}{1.0}
\caption{Standard deviations over 20 seeds for all tuned models on all algorithms. Each value in the row ``Over All Runs'' is not an average of variances for each algorithm, but rather the variance across all runs.
}
\label{tab:clrs_meandev}

\small

\begin{center}
\begin{tabular}{lrrrrrrr}\hline
{\textbf{Algorithm}} & {\textbf{Deep Sets}} & {\textbf{GAT-v1}} & {\textbf{GAT-v2}} & {\textbf{MPNN}} & {\textbf{PGN-u}} & {\textbf{PGN-m}} & {\textbf{RT (Ours)}} \\ \hline

Activity Selector & 1.7\% & 1.4\% & 2.1\% &  1.3\% & 2.6\% & 1.9\% & 2.7\% \\

Articulation Points & 6.0\% & 8.0\% & 6.3\% & 6.1\% & 3.5\% & 7.5\% & 14.6\%  \\ 

Bellman-Ford & 2.4\% & 1.5\% & 1.6\% & 1.9\% & 0.9\% & 1.3\% & 1.5\% \\

BFS & 1.0\% & 0.4\% & 0.4\% & 0.2\% & 0.3\% & 0.5\% & 0.7\% \\

Binary Search & 3.8\% & 5.8\% & 7.9\% & 5.0\% & 10.4\% & 3.7\% & 6.7\% \\

Bridges & 4.8\% & 6.4\% & 8.0\% & 17.8\% & 11.0\% & 7.8\% & 11.8\% \\

Bubble Sort & 3.1\% & 2.5\% & 1.3\% & 5.0\% & 1.9\% & 0.2\% & 13.0\% \\

DAG Shortest Paths & 3.8\% & 2.3\% & 2.2\% & 1.6\% & 3.2\% & 0.9\% & 1.6\% \\

DFS & 1.9\% & 2.1\% & 1.7\% & 2.7\% & 1.5\% & 1.5\% & 10.5\% \\

Dijkstra & 3.6\% & 9.5\% & 7.1\% & 4.3\% & 11.0\% & 3.4\% & 5.8\% \\

Find Max. Subarray & 1.5\% & 2.4\% & 2.1\% & 2.9\% & 1.6\% & 15.0\% & 3.7\% \\

Floyd-Warshall & 4.2\% & 6.0\% & 3.7\% & 4.7\% & 4.8\% & 2.5\% & 7.6\% \\

Graham Scan & 3.3\% & 6.6\% & 5.3\% & 1.4\% & 5.7\% & 6.0\% & 7.4\% \\

Heapsort & 8.8\% & 3.8\% & 4.1\% & 7.1\% & 8.9\% & 0.4\% & 14.8\% \\

Insertion Sort & 1.5\% & 14.8\% & 11.6\% & 9.1\% & 11.2\% & 0.2\% & 9.0\% \\

Jarvis' March & 8.6\% & 2.6\% & 3.9\% & 29.3\% & 2.5\% & 5.8\% & 2.2\% \\

KMP Matcher & 1.0\% & 0.6\% & 0.6\% & 1.1\% & 0.7\% & 0.4\% & 0.1\% \\

LCS Length & 6.5\% & 7.2\% & 4.6\% & 2.7\% & 7.8\% & 3.5\% & 4.1\% \\

Matrix Chain Order & 2.8\% & 5.4\% & 5.1\% & 2.3\% & 0.9\% & 1.1\% & 1.2\% \\

Minimum & 5.8\% & 2.6\% & 10.4\% & 3.4\% & 3.4\% & 27.3\% & 2.0\% \\

MST-Kruskal & 4.9\% & 2.9\% & 4.2\% & 5.9\% & 6.1\% & 4.3\% & 11.8\% \\

MST-Prim & 11.4\% & 9.2\% & 10.7\% & 5.1\% & 4.6\% & 7.2\% & 7.9\% \\

Na\"{i}ve String Match & 0.3\% & 0.7\% & 0.5\% & 1.7\% & 0.8\% & 0.7\% & 32.3\% \\

Optimal BST & 6.0\% & 5.7\% & 5.1\% & 3.5\% & 2.0\% & 2.0\% & 2.6\% \\

Quickselect & 2.4\% & 0.5\% & 0.4\% & 1.6\% & 0.9\% & 1.0\% & 17.3\% \\

Quicksort & 4.3\% & 1.3\% & 0.8\% & 3.9\% & 1.9\% & 0.2\% & 13.2\% \\

Segments Intersect & 0.7\% & 0.5\% & 0.7\% & 2.5\% & 0.8\% & 0.8\% & 2.6\% \\

SCC & 5.9\% & 8.8\% & 9.9\% & 5.2\% & 6.0\% & 6.9\% & 15.2\% \\

Task Scheduling & 0.5\% & 0.4\% & 0.8\% & 0.9\% & 0.5\% & 0.5\% & 1.8\% \\

Topological Sort & 10.3\% & 12.7\% & 17.8\% & 5.8\% & 6.5\% & 10.1\% & 17.5\% \\

\hline

Over All Runs & 29.3\% & 32.0\% & 31.3\% & 34.6\% & 33.1\% & 35.0\% & 29.6\% \\

\hline

\end{tabular}
\end{center}
\end{table*}

\begin{table*}[b]
\renewcommand{\arraystretch}{1.0}
\caption{Average test scores of all tuned models on all algorithms.
}
\label{tab:clrs_meanacc}

\small

\begin{center}
\begin{adjustbox}{width=1.1\linewidth, center}
\begin{tabular}{lrrrrrrrr}\hline
{\textbf{Algorithm}} & {\textbf{Deep Sets}} & {\textbf{GAT-v1}} & {\textbf{GAT-v2}} & {\textbf{MPNN}} & {\textbf{PGN-u}} & {\textbf{PGN-m}} & {\textbf{PGN-c}} & {\textbf{RT (Ours)}} \\ \hline

Activity Selector & 72.22\% & 68.89\% & 67.45\% & {\textbf{95.45\%}} & 67.05\% & 69.83\% & 69.83\% & 87.72\% \\

Articulation Points & 38.50\% & 31.46\% & 31.96\% & 46.21\% & 46.87\% & {\textbf{49.73\%}} & 49.73\% & 34.15\%  \\ 

Bellman-Ford & 51.00\% & 93.10\% & 93.75\% & 95.42\% & {\textbf{95.83\%}} & 95.43\% & 95.83\% & 94.24\% \\

BFS & 98.14\% & 99.76\% & 99.49\% & {\textbf{99.78\%}} & 99.71\% & 99.27\% & 99.71\% & 99.14\% \\

Binary Search & 55.77\% & 17.21\% & 31.11\% & 38.00\% & 61.71\% & {\textbf{87.03\%}} & 87.03\% & 81.48\% \\

Bridges & 36.20\% & 22.90\% & 24.07\% & {\textbf{61.29\%}} & 44.72\% & 57.15\% & 57.15\% & 37.88\% \\

Bubble Sort & {\textbf{65.03\%}} & 9.69\% & 8.46\% & 13.14\% & 7.54\% & 2.05\% & 7.54\% & 38.22\% \\

DAG Shortest Paths & 77.98\% & 86.44\% & 89.35\% & 97.33\% & 96.23\% & {\textbf{98.16\%}} & 98.16\% & 96.61\% \\

DFS & 7.62\% & 11.71\% & 12.08\% & 13.85\% & 10.78\% & 8.88\% & 10.78\% & {\textbf{39.23\%}} \\

Dijkstra & 44.51\% & 61.42\% & 68.10\% & {\textbf{92.18\%}} & 78.45\% & 90.02\% & 90.02\% & 91.20\% \\

Find Max. Subarray & 12.29\% & 15.19\% & 14.80\% & 16.14\% & 16.71\% & 51.30\% & 51.30\% & {\textbf{66.52\%}} \\

Floyd-Warshall & 6.24\% & 30.53\% & 37.58\% & 28.81\% & 27.60\% & {\textbf{39.24\%}} & 39.24\% & 31.59\% \\

Graham Scan & 64.48\% & 65.05\% & 67.07\% & {\textbf{95.31\%}} & 67.89\% & 64.66\% & 67.89\% & 74.15\% \\

Heapsort & {\textbf{72.39\%}} & 9.90\% & 7.65\% & 27.08\% & 17.88\% & 1.60\% & 17.88\% & 32.96\% \\

Insertion Sort & 72.88\% & 61.00\% & 50.19\% & 50.30\% & 82.42\% & 3.64\% & 82.42\% & {\textbf{89.43\%}} \\

Jarvis' March & 46.01\% & 55.40\% & 50.23\% & 59.31\% & 49.54\% & 49.57\% & 49.57\% & {\textbf{94.57\%}} \\

KMP Matcher & {\textbf{3.71\%}} & 1.03\% & 1.32\% & 2.35\% & 1.75\% & 0.44\% & 1.75\% & 0.03\% \\

LCS Length & 55.85\% & 48.14\% & 48.30\% & 54.42\% & 53.85\% & 56.00\% & 56.00\% & {\textbf{83.32\%}} \\

Matrix Chain Order & 81.65\% & 81.59\% & 65.58\% & 85.53\% & 86.65\% & 86.47\% & 86.65\% & {\textbf{91.89\%}} \\

Minimum & 91.69\% & 86.87\% & 80.60\% & 90.73\% & 89.22\% & 70.69\% & 89.22\% & {\textbf{95.28\%}} \\

MST-Kruskal & 69.07\% & 67.26\% & 68.60\% & {\textbf{70.84\%}} & 63.47\% & 69.35\% & 69.35\% & 64.91\% \\

MST-Prim & 35.70\% & 61.83\% & 68.72\% & {\textbf{86.43\%}} & 81.59\% & 83.61\% & 83.61\% & 85.77\% \\

Na\"{i}ve String Match & 2.12\% & 1.70\% & 1.81\% & 1.83\% & 1.46\% & 1.90\% & 1.90\% & {\textbf{65.01\%}} \\

Optimal BST & 67.37\% & 61.91\% & 63.78\% & 66.48\% & 65.19\% & 70.74\% & 70.74\% & {\textbf{74.40\%}} \\

Quickselect & 5.51\% & 2.03\% & 2.40\% & 3.07\% & 2.00\% & 4.92\% & 4.92\% & {\textbf{19.18\%}} \\

Quicksort & {\textbf{65.26\%}} & 4.41\% & 5.00\% & 17.95\% & 7.90\% & 2.63\% & 7.90\% & 39.42\% \\

Segments Intersect & 85.94\% & 86.36\% & 86.10\% & {\textbf{94.46\%}} & 85.89\% & 85.66\% & 85.89\% & 84.94\% \\

SCC & 21.13\% & 23.66\% & \textbf{29.47\%} & 12.58\% & 12.97\% & 17.20\% & 17.20\% & 28.59\% \\

Task Scheduling & 83.45\% & 82.61\% & 82.61\% & 83.36\% & 83.54\% & {\textbf{83.62\%}} & 83.62\% & 82.93\% \\

Topological Sort & 19.02\% & 43.47\% & 43.23\% & 54.93\% & 51.71\% & 64.28\% & 64.28\% & {\textbf{80.62\%}} \\

\hline

Average & 50.29\% & 46.42\% & 46.70\% & 55.15\% & 51.94\% & 52.17\% & 56.57\% & {\textbf{66.18\%}} \\

 \hline
\end{tabular}
\end{adjustbox}
\end{center}
\end{table*}

\end{document}